\documentclass[journal]{IEEEtran}
\usepackage{cite}
\usepackage{graphicx}
\usepackage{amsmath}
\usepackage{amssymb}
\usepackage{xfrac}
\usepackage{gensymb}
\usepackage{xfrac}
\usepackage{algorithmic}
\usepackage{array}
\usepackage[caption=false,font=footnotesize]{subfig}
\usepackage{url}
\usepackage{comment}
\usepackage{xcolor}
\usepackage{enumitem}
\usepackage{multirow}
\usepackage{bm}

\usepackage{acro}
\providecommand\tooltip[2]{[PDF:no. of 1][tooltip:no. of 2]}
\acsetup{tooltip=true}

\DeclareAcronym{NN}{
  short        = {NN},
  long         = {neural network},
  tooltip      = {Neural Network}
}
\DeclareAcronym{MLP}{
  short        = {MLP},
  long         = {Multilayer Perceptron},
  tooltip      = {Multilayer Perceptron}
}
\DeclareAcronym{CNN}{
  short        = {CNN},
  long         = {Convolutional Neural Network},
  tooltip      = {Convolutional Neural Network}
}
\DeclareAcronym{FC}{
  short        = {FC},
  long         = {Fully-connected},
  tooltip      = {Fully-connected}
}
\DeclareAcronym{SC}{
  short        = {SC},
  long         = {Sparsely-connected},
  tooltip      = {Sparsely-connected}
}
\DeclareAcronym{PCA}{
  short        = {PCA},
  long         = {Principal Component Analysis},
  tooltip      = {Principal Component Analysis}
}
\DeclareAcronym{FF}{
  short        = {FF},
  long         = {Feedforward},
  tooltip      = {Feedforward}
}
\DeclareAcronym{BP}{
  short        = {BP},
  long         = {Backpropagation},
  tooltip      = {Backpropagation}
}
\DeclareAcronym{UP}{
  short        = {UP},
  long         = {Update of Trainable Parameters},
  tooltip      = {Update of Trainable Parameters}
}
\DeclareAcronym{MFCC}{
  short        = {MFCC},
  long         = {Mel-frequency Cepstral Coefficient},
  tooltip      = {Mel-frequency Cepstral Coefficient}
}
\DeclareAcronym{TPC}{
  short        = {TPC},
  long         = {Test Prediction Comparison},
  tooltip      = {Test Prediction Comparison}
}
\DeclareAcronym{CI}{
  short        = {CI},
  long         = {Confidence Interval},
  tooltip      = {Confidence Interval}
}

\DeclareAcronym{GPU}{
  short        = {GPU},
  long         = {Graphical Processing Unit},
  tooltip      = {Graphical Processing Unit}
}

\DeclareAcronym{CPU}{
  short        = {CPU},
  long         = {Central Processing Unit},
  tooltip      = {Central Processing Unit}
}

\DeclareAcronym{FPGA}{
  short        = {FPGA},
  long         = {Field Programmable Gate Array},
  tooltip      = {Field Programmable Gate Array}
}

\DeclareAcronym{LDPC}{
  short        = {LDPC},
  long         = {Low Density Parity Check},
  tooltip      = {Low Density Parity Check}
}

\DeclareAcronym{GCD}{
  short        = {$\mathrm{gcd}$},
  long         = {Greatest Common Divisor},
  tooltip      = {Greatest Common Divisor}
}

\DeclareAcronym{LSS}{
  short        = {$\mathrm{LSS}$},
  long         = {Learning Structured Sparsity},
  tooltip      = {Learning Structured Sparsity}
}

\DeclareAcronym{ASR}{
  short        = {ASR},
  long         = {Automatic Speech Recognition},
  tooltip      = {Automatic Speech Recognition}
}

\newcommand{\eg} {\emph{e.g.,~}}
\newcommand{\ie} {\emph{i.e.,~}}

\renewcommand{\theenumii}{\theenumi.\arabic{enumii}.}

\newcommand{\indeg}{d^{\mathrm{in}}}
\newcommand{\outdeg}{d^{\mathrm{out}}}
\newcommand{\outdegnet}{\bm{d}^{\mathrm{out}}_\mathrm{net}}
\newcommand{\Nnet}{\bm{N}_{\mathrm{net}}}
\newcommand{\countNmi}{S_{M_i}}
\newcommand{\countNci}{S_{C_i}}
\newcommand{\znet}{\bm{z}_{\mathrm{net}}}
\newcommand{\rhonet}{\rho_{\mathrm{net}}}

\newcommand{\serifmod}{\textsf{\,\%\,}}

\begin{document}
%\bstctlcite{IEEEexample:BSTcontrol}
\title{Pre-Defined Sparse Neural Networks\\with Hardware Acceleration}

\author{Sourya~Dey, %~\IEEEmembership{Member,~IEEE,}
        Kuan-Wen~Huang, %~\IEEEmembership{Fellow,~OSA,}
        Peter~A.~Beerel,~\IEEEmembership{Senior~Member,~IEEE,}
        and~Keith~M.~Chugg,~\IEEEmembership{Fellow,~IEEE}% <-this % stops a space
\thanks{The authors are with the Ming Hsieh Department
of Electrical Engineering, University of Southern California, Los Angeles,
CA, 90089 USA e-mail: \{souryade, kuanwenh, pabeerel, chugg\}@usc.edu}% <-this % stops a space
\thanks{Manuscript submitted December 3, 2018.}
\thanks{This work is partly supported by NSF, Software and Hardware Foundations,  Grant 1763747.}}
%\thanks{This work was presented in part at ICANN 2017 \cite{Dey2017_ICANN}, Asilomar 2017 \cite{Dey2017_Asilomar}, ITA 2018 \cite{Dey2018_ITA}, and will be presented at ReConFig 2018 \cite{Dey2018_Reconfig}.}



\maketitle

\begin{abstract}
Neural networks have proven to be extremely powerful tools for modern artificial intelligence applications, but computational and storage complexity remain limiting factors.  This paper presents two compatible contributions towards reducing the time, energy, computational, and storage complexities associated with multilayer perceptrons.  Pre-defined sparsity is proposed to reduce the complexity during both training and inference, regardless of the implementation platform.  Our results show that storage and computational complexity can be reduced by factors greater than 5X without significant performance loss.  The second contribution is an architecture for hardware acceleration that is compatible with pre-defined sparsity.  This architecture supports both training and inference modes and is flexible in the sense that it is not tied to a specific number of neurons.  For example, this flexibility implies that various sized neural networks can be supported on various sized \ac{FPGA}s.
\end{abstract}

\begin{IEEEkeywords}
Machine learning, Neural network, Multilayer perceptron, Sparsity, Hardware Acceleration
\end{IEEEkeywords}

% \fi
\IEEEpeerreviewmaketitle

\section{Introduction}\label{sec-intro}
\IEEEPARstart{N}{eural} networks  are critical drivers of new technologies such as computer vision, speech recognition, and autonomous systems. As more data have become available, the size and complexity of \ac{NN}s has risen sharply  with modern \ac{NN}s containing millions or even billions of  trainable parameters  \cite{Krizhevsky2012_alexnet,Coates2013}.  These massive \ac{NN}s come with the cost of large computational and storage demands.  The current state of the art is to train large \ac{NN}s on \ac{GPU}s in the cloud -- a process that can take days to weeks even on powerful \ac{GPU}s \cite{Krizhevsky2012_alexnet,Coates2013,Han2015_learning} or similar programmable processors with multiply-accumulate accelerators \cite{Jouppi2017_TPU}.  Once trained, the model can be used for inference which is less computationally intensive and is typically performed on more general purpose processors (\ie \ac{CPU}s).  It is increasingly desirable to run inference, and even some re-training, on embedded processors which have limited resources for computation and storage.
 In this regard, model reduction has been identified as a key to \ac{NN} acceleration by several prominent researchers \cite{szegedy15deeper}.
 This is generally performed post-training to reduce the memory requirements to store the model for inference -- \eg methods for quantization, compression, and grouping parameters \cite{Gong2014,Chen2015,Han2015_DC,Han2016}.

Decreasing the time, computation, storage,  and energy costs for training and inference is therefore a highly relevant goal.  In this paper we present two compatible methods towards this end goal: (i) a method for introducing sparsity in the connection patterns of  \ac{NN}s, and (ii) a flexible hardware architecture that is compatible with training and inference-only operation and supports the proposed sparse \ac{NN}s.  Our approach to sparsifying a \ac{NN} is extremely simple and results in a large reduction in storage and computational complexity both in training and inference modes.  Moreover, this method is not tied to the hardware acceleration and provides the same benefits for training and inference in software under the current paradigm.  The hardware architecture is massively parallel, but not tightly coupled to a specific \ac{NN} architecture (\ie not tied to the number of nodes in a layer).  Instead, the architecture allows for maximum throughput for a given amount of circuit resources.  

Our approach to making a \ac{NN} sparse is to specify a sparse set of neuron connections prior to training and to hold this pattern fixed throughout training and inference. We refer to this method of simply excluding some fixed set of connections in the \ac{NN} as \emph{pre-defined sparsity}.  
There are several methods in the literature related to sparse \ac{NN}s, but most do not reduce the computation and storage complexity associated with training, which is a primary goal of this work.  One related concept is drop-out \cite{Srivastava14} where selected edges in the \ac{NN} are not processed during some steps of the training process, but the final result is a \ac{FC} \ac{NN} for inference.  Another set of approaches target producing a sparse \ac{NN} for inference, but use \ac{FC} \ac{NN}s during training.  Among these are pruning and trimming methods  that post-process the trained \ac{NN} to produce a sparse \ac{NN} for inference mode \cite{Albericio2016,Reagen2016,nettrim-plus}.  As mentioned above, other methods have been proposed for reducing the complexity of performing inference on a trained \ac{FC} \ac{NN} such as quantization, compression, and grouping parameters \cite{Gong2014,Chen2015,Han2015_DC,Han2016}.  Other  research has suggested a method of learning sparsity during training that begins training a fully-connected \ac{NN} and uses a cost regularizer that promotes sparsity in the trained model \cite{NIPS2016_6504}.  Note that all of these methods do not substantially reduce the complexity of training and instead target inference models that have lower complexity.  One method aimed at reducing both training and inference complexity is using \ac{NN}s with structured, but not sparse, weight matrices \cite{Sindhwani2015,Wang2018}. Finally, we note that several authors have very recently proposed pre-defined sparse \ac{NN}s \cite{Bourely2017,Prabhu2017,Mocanu2018} independently of our published work \cite{Dey2017_ICANN,Dey2017_Asilomar,Dey2018_ITA}.  

Motivated by the fact that specialized hardware is typically faster and more energy efficient than \ac{GPU}s and \ac{CPU}s, there exists a large body of literature in \ac{NN} hardware acceleration.  The vast majority of this addresses only inference given a trained model \cite{Chen2017,Han2016,Yufei2017,Cambricon,Suda2016}, with few addressing hardware accelerated training \cite{Chen:2014}. The work of \cite{Chen:2014}, for example, targets a specific size \ac{NN} -- \ie the logic and memory architecture is tied to the number of neurons in a layer.

We propose an architecture that supports training, but can be simplified for inference-only mode, and is flexible to the \ac{NN} size.  This is particularly attractive for \ac{FPGA} implementations.  Specifically, the proposed architecture produces the maximum throughput on a given \ac{FPGA} for a given \ac{NN} and can therefore support various sized \ac{NN}s on various sized \ac{FPGA}s.   This is accomplished by an \emph{edge-based} processing architecture that can process $z$ edges in a given layer in parallel (\ie we refer to $z$ as the \emph{degree of parallelism}).   A given FPGA can support some largest value of $z$, and \ac{NN}s with more edges will simply take more clock cycles to process.\footnote{We use the terms the terms `connection' and  `edge' interchangeably, as we do with `node' and `neuron'. Also, the term `cycle' will mean `clock cycle', unless otherwise stated.}  

Our edge-based architecture is inspired by architectures proposed for iterative decoding of modern sparse-graph-based error correction codes (\ie Turbo and \ac{LDPC} codes) (cf., \cite{masera1999vlsi,brack2007low }).  In particular, for a given processing task, there are $z$ logic units to perform the task and $z$ memories to store the quantities associated with the task.  A challenge with this architecture, shared between the decoding and \ac{NN} applications, is that, in order to achieve high-throughput without memory duplication,  the parallel memories must be accessed in two manners: natural order and interleaved order.  In natural order, each computation unit is associated with one memory and accesses the elements of that memory sequentially. For interleaved order access, the $z$ computational units must access the memories such that no memory is accessed more than once in a cycle.  Such an addressing pattern is called \emph{clash-free}, and this property ensures that no memory contention occurs so that no stalls or wait states are required.  For modern codes, the clash-free property of the memories is ensured by defining clash-free interleavers (\ie permutations) \cite{crozier2001high}, or clash-free parity check matrices \cite{brack2007low}.  In the context of \ac{NN}s, this clash-free property is tied to the connection patterns between layers of neurons.

In addition to $z$ degrees of parallelism in edge processing in a given layer, our architecture is pipelined across layers.  Thus, there is a degree of parallelism associated with each layer (\ie $z_i$ for layer $i$) selected to set the number of cycles required to process a layer to a constant -- \ie larger layers have larger $z$ so that the computation time of all layers is the same.  For an $(L+1)$-layer \ac{NN} there are $L$ pipeline stages so that a given \ac{NN} input is processed in the time it takes to complete the processing of the edges in a single layer.  Furthermore, the three operations associated with training -- \ac{FF}, \ac{BP}, and \ac{UP} -- are performed in parallel.  The architecture may be simplified to perform only inference by eliminating the logic and memory associated with \ac{BP} and \ac{UP}.  Furthermore, while the architecture supports the reduced sparse complexity \ac{NN}s, it is also compatible with traditional \ac{FC} networks.  Interestingly, very recent work proposed pipelining across layers for an inference-only accelerator \cite{Zhou2017a}, as well as a scalable edge-based architecture for training \cite{Zhou2017} independently of our published work \cite{Dey2017_ICANN,Dey2017_Asilomar}.  Neither of these other recent works, however, takes advantage of pre-defined sparsity in the network.

In Section \ref{sec-spds} we provide motivation for and simple examples of the effectiveness of pre-defined sparsity.  In Section \ref{sec-hardware}  the hardware architecture is described in detail, including defining a class of simple clash-free connection patterns with low address generation complexity.  Section  \ref{sec-trends} contains a detailed simulation study of pre-defined sparsity in \ac{NN}s based on four different classification datasets -- MNIST handwritten digits \cite{mnist-dataset}, Reuters news articles \cite{reuters-rcv1v2-dataset}, TIMIT speech corpus \cite{timit-dataset}, and CIFAR-100 images \cite{cifar-dataset}.  We identify a set of trends or design guidelines in this section as well.  This section also demonstrates that the simple, hardware-compatible clash-free connection patterns provide performance on-par or better than that of randomly connected sparse patterns.  Finally, in Section \ref{sec-othersparsity} we consider the issue of whether pre-defining the structured sparse patterns causes a significant performance loss relative to other sparse methods having similar amount of parameters. We find that there is no significant performance degradation and therefore our hardware architecture can provide training and inference performance commensurate with state-of-the art sparsity methods.

\section{Structured Pre-Defined Sparsity}\label{sec-spds}

\subsection{Definitions, Notation, and Background}\label{sec-spds-basics}
An $(L+1)$-layer \ac{MLP} has $N_i$ nodes in the $i^{\mathrm{th}}$ layer, described collectively by the \emph{neuronal configuration} $\Nnet = \left(N_0,N_1,\cdots,N_L\right)$, where layer 0 is the input layer. We use the convention that layer $i$ is to the `right' of layer $i-1$. There are $L$ \emph{junctions} between layers, with junction $i$ connecting the $N_{i-1}$ nodes of its left layer $i-1$ with the $N_{i}$ nodes of its right layer $i$.

We define pre-defined sparsity as simply not having all $N_{i-1} N_i$ edges present in junction $i$.  Furthermore, we define \emph{structured} pre-defined sparsity so that for a given junction $i$, each node in its left layer has fixed out-degree -- \ie $\outdeg_i$ connections to its right layer, and each node in its right layer has fixed in-degree -- \ie $\indeg_i$ connections from its left layer.  \ac{FC} \ac{NN}s have $\outdeg_i = N_i$ and $\indeg_i = N_{i-1}$ with $N_{i-1}N_i$ edges in the $i^{\mathrm{th}}$ junction, while a sparse \ac{NN} has at least one junction with less than this number of edges.  The number of edges (or weights) in junction $i$ is given by $|\bm{W}_i| = N_{i-1} \outdeg_{i} = N_{i} \indeg_{i}$.  The density of junction $i$ is measured relative to \ac{FC} and denoted as $\rho_i = |\bm{W}_i| / (N_{i-1} N_i)$.  The structured constraint implies that the number of possible $\rho_i$ values is equal to the \ac{GCD} of $N_{i-1}$ and $N_i$, as shown in Appendix \ref{appendix-spds-constraints}. The overall density is
\begin{IEEEeqnarray}{c}\label{eq-overalldensity}
\rhonet =  \frac{ \sum_{i=1}^{L} |\bm{W}_i|} {\sum_{i=1}^{L} N_{i-1}{N_i}}
\end{IEEEeqnarray}
Thus, specifying $\Nnet$ and the \emph{out-degree configuration} $\outdegnet = ( \outdeg_1, \cdots , \outdeg_L )$ determines the density of each junction and the overall density. 

We will also consider \emph{random pre-defined sparsity}, where connections are distributed randomly given preset $\rho_i$ values without constraints on in- and out-degrees.  In Sec. \ref{sec-predefinedsparsity-comparison} we show that random pre-defined sparsity is undesirable at low densities because it may result in unconnected neurons.

The standard equations for \ac{FC} \ac{NN}s are well-known \cite{Goodfellow-et-al-2016}. For a \ac{NN} using structured pre-defined sparsity, only the  weights corresponding to connected edges are stored in memory and used in computation. This leads to the modified equations \eqref{eq-ff}--\eqref{eq-up}, where subscripts denote layer/junction numbers, single superscripts denote neurons in a layer, and double superscripts denote (right neuron, left neuron) in a junction.  The \ac{FF} processing proceeds left-to-right and computes the activations  $\bm{a}_i$ and associated derivatives $\dot{\bm{a}}_i$ for each layer by applying an activation function $\text{act}(\cdot)$ to a linear combination of biases $\bm{b}_i$, junction weights $\bm{W}_i$ and preceding layer activations $\bm{a}_{i-1}$
        \begin{IEEEeqnarray}{rCl}\label{eq-ff}
        h_i^{(j)} &=& \sum _{f=1}^{\indeg_i} { W_{i}^{(j,k_f)}a_{i-1}^{(k_f)} + b_i^{(j)} } \IEEEyesnumber \IEEEyessubnumber \label{eq-ff-a} \\
        a_i^{(j)} &=& \text{act} \left( h_i^{(j)} \right) \IEEEyessubnumber \label{eq-ff-b} \\
        \dot {a}_i^{(j)} &=& \frac{da_i^{(j)}}{dh_i^{(j)}} = \dot{\text{act}} \left( h_i^{(j)} \right) \IEEEyessubnumber \label{eq-ff-c}
        \end{IEEEeqnarray}
Note that \eqref{eq-ff-c} is used in training, but is not required in inference mode.  
The \ac{BP} computation is done only in training and computes a sequence of error values from right-to-left  
      \begin{IEEEeqnarray}{rCl}\label{eq-bp}
        \delta_L^{(j)} &=& \frac{\partial l^{(j)} \left(a_L^{(j)},y^{(j)}\right)}{\partial h_L^{(j)}} \IEEEyesnumber \IEEEyessubnumber \label{eq-bp-a} \\
        \delta_i^{(j)} &=& \dot {a}_i^{(j)} \left( \sum _{f=1}^{\outdeg_i} { W_{i+1}^{(k_f,j)}{\delta}_{i+1}^{(k_f)} } \right) \IEEEyessubnumber \label{eq-bp-b}
        \end{IEEEeqnarray}
where $l^{(j)} \left(a_L^{(j)},y^{(j)}\right)$ is the  $j^{\mathrm{th}}$ component of the loss function.  Finally, stochastic gradient \ac{UP} is given by 
        \begin{IEEEeqnarray}{rCl}\label{eq-up}
        b_i^{(j)} &\leftarrow& b_i^{(j)} - \eta {\delta}_i^{(j)} \IEEEyesnumber \IEEEyessubnumber \label{eq-up-a} \\
        W_i^{(j,k)} &\leftarrow& W_i^{(j,k)} - \eta a_{i-1}^{(k)}{\delta}_i^{(j)} \IEEEyessubnumber \label{eq-up-b}
        \end{IEEEeqnarray}   
where $\eta$ is the learning rate.
The parameters on left-hand-side of \eqref{eq-ff}--\eqref{eq-up} will be referred to as the \emph{network parameters}, with the weights and biases being the  \emph{trainable parameters}.

\subsection{Motivation and Preliminary Examples}\label{sec-spds-expapp}
\begin{figure}[!t]
\centering
\includegraphics[width = 1.0\linewidth]{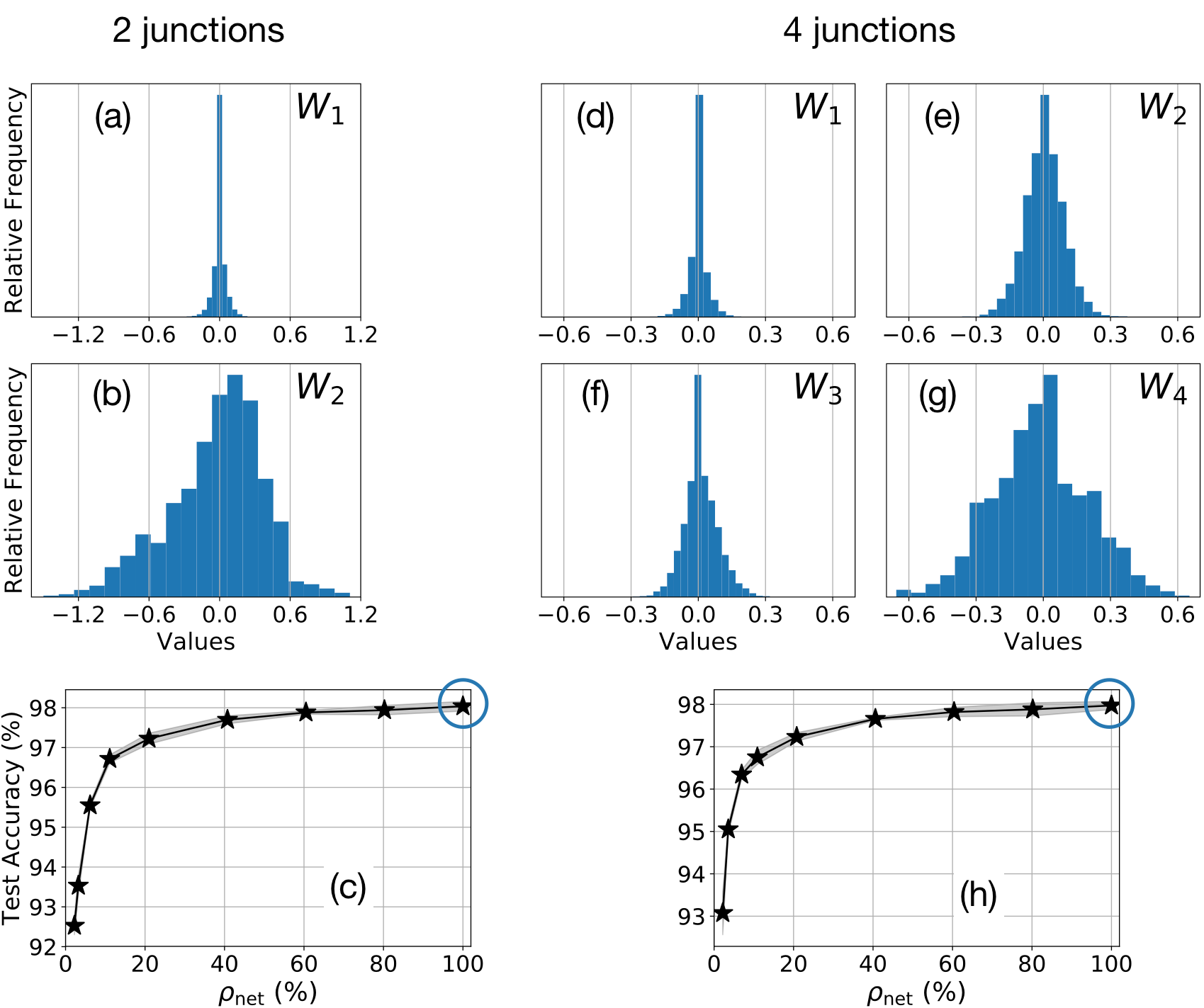}
\caption{Histograms of weight values in different junctions for \ac{FC} \ac{NN}s trained on MNIST for 50 epochs, with (a-b) $\Nnet = (800,100,10)$, and (d-g) $\Nnet = (800,100,100,100,10)$.  Test accuracy shown in (c,h) for different \ac{NN}s with same $\Nnet$ and varying $\rhonet$. The overall density $\rhonet$ is set by reducing $\rho_1$ since junction 1 has more weights close to zero in the \ac{FC} cases (circled).}
\label{fig-pdsexample}
\end{figure}

\begin{figure*}[!t]
\centering
\includegraphics[width = 0.9\linewidth]{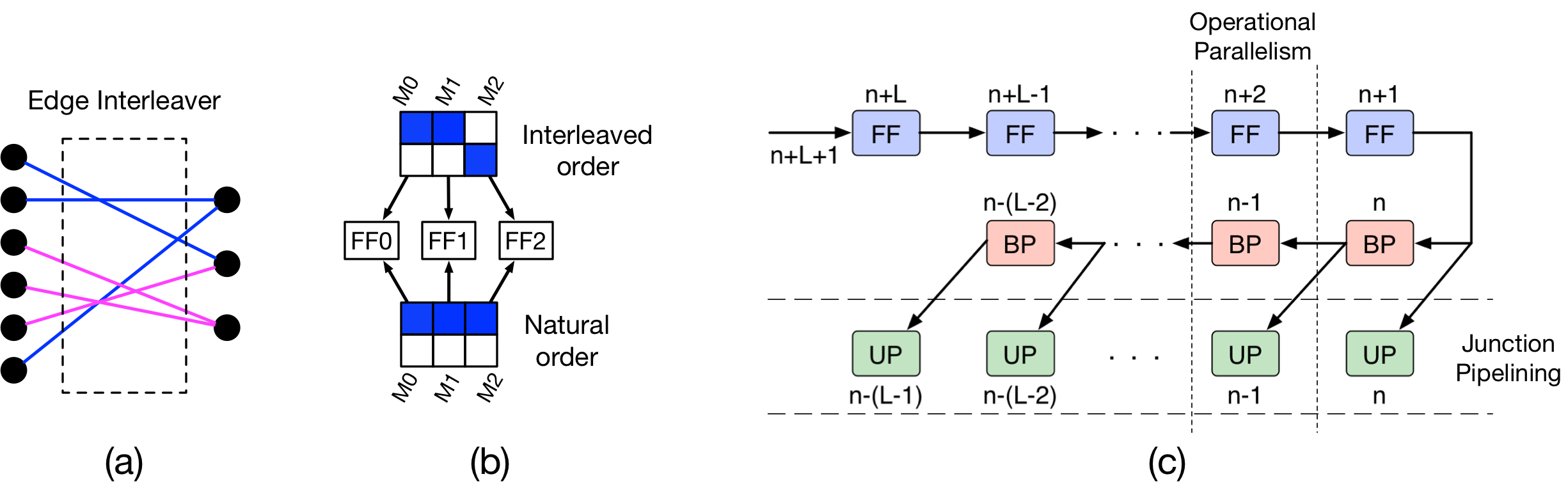}
\caption{(a) Processing $z_i = 3$ edges in each cycle (blue in cycle 0, pink in cycle 1) for some junction $i$% of the single junction network from Fig. \ref{fig-pdsbasic}(c-d)
. (b) Accessing $z_i = 3$ memories -- M0, M1 and M2 shown as columns -- from two separate banks, one in natural order (same address from each memory), the other in interleaved order. Clash-freedom is achieved by accessing only one element from each memory. The accessed values are fed to $z_i = 3$ processors to perform \ac{FF} simultaneously. (c) Operational parallelism in each junction (vertical dotted lines denote processing for one junction), and junction pipelining of each operation across junctions (horizontal dashed lines) in a multi-junction \ac{NN}. Subfigure (c) is modified from our previous conference publication \cite[Fig. 2(c)]{Dey2017_ICANN}}
\label{fig-hardwareoverview}
\end{figure*}

Pre-defined sparsity can be motivated by inspecting the histogram for trained weights in a \ac{FC} \ac{NN}. There have been previous efforts to study such statistics \cite{Han2015_learning,Yosinski2012}, however, not for individual junctions. Fig. \ref{fig-pdsexample} shows weight histograms for each junction in both a 2-junction and 4-junction \ac{FC} \ac{NN} trained on the MNIST dataset. Note that many of the weights are zero or near-zero after training, especially in the earlier junctions. This motivates the idea that some weights in these layers could be set to zero (\ie the edges excluded). Even with this intuition, it is unclear that one can pre-define a set of weights to be zero and let the \ac{NN} learn around this pre-defined sparsity constraint.  Fig. \ref{fig-pdsexample}(c) and (h) show that, in fact, this is the case -- \ie this shows classification accuracy as a function of the overall density $\rhonet$ for structured pre-defined sparsity.  Since the computational and storage complexity is directly proportional to the number of edges in the \ac{NN}, operating at an overall density of, for example,  50\% results in a 2X reduction in complexity both during training and inference.  Detailed numerical experiments in Section \ref{sec-trends} build on these simple examples.  However, before we proceed to those results, it is important to consider a hardware architecture that can support structured pre-defined sparsity and consider the additional clash-free constraints placed on the connection patterns so that these can be considered in the studies in Section \ref{sec-trends}.  

\section{Hardware Architecture}\label{sec-hardware}
In this section we describe the proposed flexible hardware architecture outlined in the Introduction.  The overall architectural view is captured by Fig. \ref{fig-hardwareoverview}: sub-figure (a) shows parallel edge processing within a junction with degree of parallelism 3, (b) shows clash-free memory access, and (c) junction pipelining and parallel processing of the three operations -- \ac{FF}, \ac{BP}, \ac{UP}.  The toy example in Fig. \ref{fig-hardwareoverview}(a)-(b) is for $N_{i-1}=6$, $N_{i}=3$, $\rho_i = 6/18 = 1/3$,  and $z_i = 3$.  Fig. \ref{fig-hardwareoverview}(a) shows that the $z_i=3$ blue edges are processed in parallel in one cycle, while the pink edges are processed in parallel during the next cycle.  Fig. \ref{fig-hardwareoverview}(b) shows how the $z_i = 3$ \ac{FF} processing logic units access the memories in natural and interleaved order.  As described in detail in Sec. \ref{sec-hardware-memoryorg}, the interleaved order access may represent reading of the activations $\{ a_{i-1}^{(j)} \}$ for $j \in \{0,1, 5 \}$ and the natural order access may correspond to writing the computed activations $\{ a_i^{(j)} \}$ for $j \in \{0,1, 2 \}$.  On the next cycle, the remaining memory locations (\ie the white cells) will be accessed.  Note that this illustrates a clash-free connection pattern since each of the $z_i = 3$ memories is accessed no more than once in each cycle -- \ie one hit per column on each access.  

The junction-based operation in Fig. \ref{fig-hardwareoverview}(b) is repeated for each junction in a pipeline. In particular, there are $L$ pipeline stages.  For example, for the \ac{FF} pipeline, while the first stage is processing input vector $n+L$ on junction 1, the second stage is processing input vector $n+L-1$ on junction 2.  The degree of parallelism for each junction is selected so that the processing time for any operation (\ac{FF}/\ac{BP}/\ac{UP}) is the same for each junction. Thus the throughput, \ie the frequency of processing input samples, is determined by the time taken to perform a single operation in a single junction.

In summary, the architecture is (i) edge-based and not tied to a specific number of nodes in a layer, (ii) flexible in that the amount of logic is determined by the degree of parallelism which trades size for speed, and (iii) fully pipelined for the parallel operations associated with \ac{NN} training.  Also note that the architecture can be specialized to perform only inference by removing the logic and memory associated with the \ac{BP} and \ac{UP} operations, and the $\dot{\bm{a}}_i$ computation in \eqref{eq-ff-c}.

A key concern when implementing \ac{NN}s on hardware is the large amount of storage required. Several characteristics regarding memory requirements guided us in developing the proposed architecture.  Firstly, since weight memories are the largest, their number should be minimized. Secondly, having a few deep memories is more efficient in terms of power and area than having many shallow memories \cite{WesteHarris}.  Thirdly, throughput should be maximized without duplicating memories, hence the need for clash-free connection patterns.

In Sec. \ref{sec-hardware-jpop}, we describe junction pipelining design which attempts to minimize weight storage resources.  The memory organization within a junction is described in  Sec. \ref{sec-hardware-memoryorg}, and is designed to minimize the number of memories for a given degree of parallelism. Finally, clash-free access conditions are developed in Sec. \ref{sec-hardware-memoryorg} and \ref{sec-hardware-clashfreedom}, and a simple method for implementing such patterns given in Sec. \ref{sec-hardware-clashfreedom}.

\subsection{Junction pipelining and Operational parallelism}\label{sec-hardware-jpop}
Our edge-based architecture is motivated by the fact that all three operations -- \ac{FF}, \ac{BP}, \ac{UP} -- use the same weight values for computation. Since $z_i$ edges are processed in parallel in a single cycle, the time taken to complete an operation in junction $i$ is $\left(C_i = \left|\bm{W}_i\right|/z_i\right)$ cycles. The \emph{degree of parallelism configuration} $\znet = \left(z_1,\cdots,z_L\right)$ is chosen to achieve $C_i = C \quad \forall\, i \in \{1,\cdots,L\}$. This allows efficient junction pipelining since each operation takes exactly $C$ cycles to be completed for each input in each junction, which we refer to as a \emph{junction cycle}.\footnote{During hardware implementation, a few extra cycles may be needed to flush the pipeline so that $C_i = \left|\bm{W}_i\right|/z_i+c_i$. These are also balanced, \ie $c_i = c \quad \forall~ i \in \{1,\cdots,L\}$, to achieve efficient pipelining. In our initial implementation \cite{Dey2018_Reconfig}, for example, $c=2$ and the junction cycle is $C=34$.} This determines throughput.

\begin{figure}[!t]
\centering
\includegraphics[width = 0.95\linewidth]{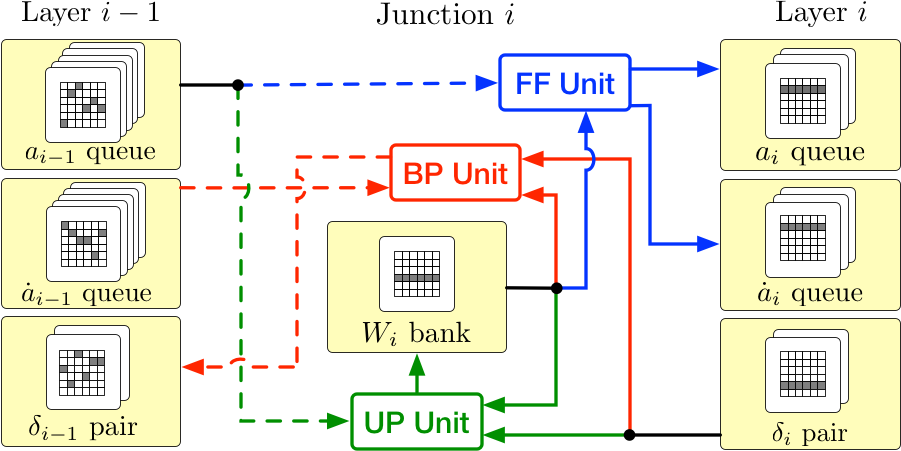}
\caption{Architecture for parallel operations  for an intermediate junction $i$ ($i \ne 1,L$) showing the three operations along with associated inputs and outputs.  Natural and interleaved order accesses are shown using solid and dashed lines, respectively. The $\bm{a}$ and $\dot{\bm{a}}$ memory banks occur as queues, the $\bm{\delta}$ memory banks as pairs, while there is a single weight memory bank. Figure modified from our previous conference publication \cite[Fig. 3]{Dey2017_ICANN}.}
\label{fig-opparall}
\end{figure}

The following is an analysis of Fig. \ref{fig-hardwareoverview}(c) in more detail for an example \ac{NN} with $L=2$. While a new training input numbered $n+3$ is getting loaded as $\bm{a}_0$, junction 1 is processing the \ac{FF} stage for the previous input $n+2$ and computing $\bm{a}_1$. Simultaneously, junction 2 is processing \ac{FF} and computing cost $\bm{\delta}_L$ via cost derivatives for input $n+1$. It is also doing \ac{BP} on input $n$ to compute $\bm{\delta}_1$, as well as updating (\ac{UP}) its parameters from the finished $\bm{\delta}_L$ computation of input $n$. Simultaneously, junction 1 is performing \ac{UP} using $\bm{\delta}_1$ from the finished \ac{BP} results of input $n-1$.\footnote{Note that \ac{BP} does not occur in the first junction because there are no $\bm{\delta}_0$ values to be computed} This results in \emph{operational parallelism} in each junction, as shown in Fig. \ref{fig-opparall}. The combined speedup is approximately a factor of $3L$ as compared to doing one operation at a time for a single input.

Notice from Fig. \ref{fig-opparall} that there is only one weight memory bank which is accessed for all three operations. However, \ac{UP} in junction $1$ needs access to $\bm{a}_0$ for input $n-1$, as per the weight update equation \eqref{eq-up-b}. This means that there need to be $2L+1 = 5$ left activation memory banks for storing $\bm{a}_0$ for inputs $n-1$ to $n+3$, \ie a queue-like structure. Similarly, \ac{UP} in junction 2 will need $2(L-1)+1 = 3$ queued banks for each of its left activation $\bm{a}_1$ and its derivative $\dot{\bm{a}}_1$ memories -- for inputs from $n$ (for which values will be read) to $n+2$ (for which values are being computed and written). There also need to be 2 banks for all $\bm{\delta}$ memories -- 1 for reading and the other for writing. Thus junction pipelining requires multiple memory banks, but only for layer parameters $\bm{a}$, $\dot{\bm{a}}$ and $\bm{\delta}$, \emph{not} for weights.\footnote{This is achieved by making the weight memory dual-port, while $\bm{a}$ and $\dot{\bm{a}}$ are single-ported memories. The $\bm{\delta}$ memories are also dual-ported due to the exact manner in which we implemented this architecture on \ac{FPGA}, refer to \cite{Dey2018_Reconfig} for full details.} The number of layer parameters is insignificant compared to the number of weights for practical networks. This is why pre-defined sparsity leads to significant storage savings, as quantified in Table \ref{table-storagecost} for the circled \ac{FC} point vs the $\rhonet = 21\%$ point from Fig. \ref{fig-pdsexample}(c). Specifically, memory requirements are reduced by 3.9X in this case.  Furthermore, the computational complexity, which is proportional to the number of weights for a \ac{MLP}, is reduced by 4.8X.  For this example, these complexity reductions come at a cost of degrading the classification accuracy from $98.0\%$ to $97.2\%$.

\begin{table}[!t]
\renewcommand{\arraystretch}{1.5}
% if using array.sty, it might be a good idea to tweak the value of \extrarowheight as needed to properly center the text within the cells
\caption{Hardware Architecture Total Storage Cost Comparison for $\Nnet = (800,100,10)$ \ac{FC} vs sparse with $\outdegnet = (20,10)$, $\rhonet = 21\%$}
\label{table-storagecost}
\centering
\begin{tabular}{|c|c|c|c|}
\hline
Parameter & Expression & Count (\ac{FC}) & Count (sparse) \\
\hline
$\bm{a}$ & $\sum_{i=0}^{L-1}{\left(2(L-i)+1\right)N_i}$ & 4300 & 4300 \\
\hline
$\dot{\bm{a}}$ & $\sum_{i=1}^{L-1}{\left(2(L-i)+1\right)N_i}$ & 300 & 300 \\
\hline
$\bm{\delta}$ & 2$\sum_{i=1}^{L}{N_i}$ & 220 & 220 \\
\hline
$\bm{b}$ & $\sum_{i=1}^{L}{N_i}$ & 110 & 110 \\
\hline
$\bm{W}$ & $\sum_{i=1}^{L}{N_i\indeg_i}$ & \textbf{81000} & \textbf{17000} \\
\hline
TOTAL & $\Sigma$ (All above) & 85930 & 21930 \\
\hline
\end{tabular}
\end{table}

%However, operational parallelism is arithmetic unit intensive. Since $z_i$ weights are read and used for all 3 operations, there need to be at least $3z_i$ multipliers per junction. A more detailed count of hardware resource utilization and implementation practicalities is given in our previous work \cite{Dey2018_Reconfig}, and is not repeated here in the interest of brevity.

\subsection{Memory organization}\label{sec-hardware-memoryorg}
\begin{figure}[!t]
\centering
\includegraphics[width = 1.0\linewidth]{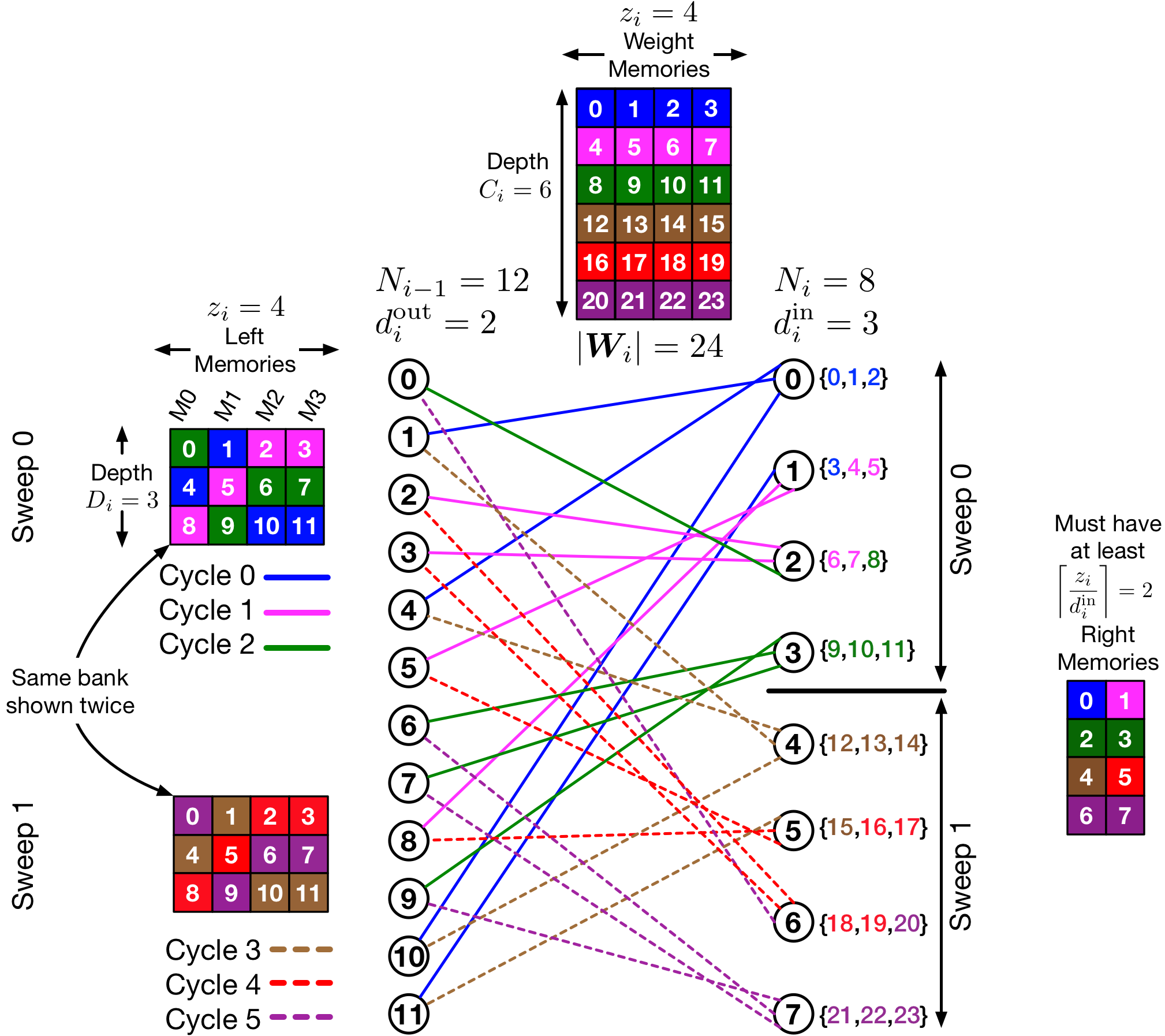}
\caption{An example of processing inside junction $i$ with $z_i=4$ memories in the weight and left banks, and $z_{i+1}=2$ memories in the right bank. The banks are represented as numerical grids, each column is a memory, and the number in each cell is the number of the edge / left neuron / right neuron whose parameter value is stored in it. Edge are sequentially numbered on the right (shown in curly braces). Four weights are read in each of the six cycles with the first three  colored blue, pink and green, respectively.  These represent sweep 0, while the next 3 (using dashed lines)  colored brown, red and purple, respectively,  represent sweep 1. Clash-freedom leads to at most one cell from each memory in each bank being accessed each cycle. Weight and right memories are accessed in natural order, while left memories are accessed in interleaved order.}
\label{fig-memaccesses}
\end{figure}

For the purposes of memory organization, edges are numbered sequentially from top to bottom on the right side of the junction. Other network parameters such as $\bm{a}$, $\dot{\bm{a}}$ and $\bm{\delta}$ are numbered according to the neuron numbers in their respective layer. Consider Fig. \ref{fig-memaccesses} as an example, where junction $i$ is flanked by $N_{i-1}=12$ left neurons with $\outdeg_i=2$ and $N_i=8$ right neurons, leading to $\left|\bm{W}_i\right| = 24$ and $\indeg_i=3$. The three weights connecting to right neuron 0 are numbered 0, 1, 2;  the next three connecting to right neuron 1 are numbered 3, 4, 5, and so on. A particular right neuron connects to some subset of left neurons of cardinality $\indeg_i$.

Each type of network parameter is stored in a bank of memories. The example in Fig. \ref{fig-memaccesses} uses $z_i = 4$, \ie 4 weights are accessed per cycle. We designed the weight memory bank to have the minimum number of memories to prevent clashes, \ie $z_i$, and their depth equals $C_i$. Weight memories are read in natural order -- 1 row per cycle (shown in same color).

Right neurons are processed sequentially due to the weight numbering. The number of right neuron parameters of a particular type needing to be accessed in a cycle is upper bounded by $\left\lceil z_i/\indeg_i \right\rceil$, which leads to $z_{i+1} \geq \left\lceil z_i/\indeg_i \right\rceil$ in order to prevent clashes in the right memory bank.\footnote{This does not limit most practical designs (see Appendix \ref{appendix-hardware-constraints}).} For \ac{FF} in Fig. \ref{fig-memaccesses} for example, cycles 0 and 1 finish computation of $a_i^{(0)}$ and $a_i^{(1)}$ respectively, while cycle 2 finishes computing both $a_i^{(2)}$ and $a_i^{(3)}$. For \ac{BP} or \ac{UP}, everything remains same except for the right memory accesses. Now $\delta_i^{(0)}$ and $\delta_i^{(1)}$ are used in cycle 0, $\delta_i^{(1)}$ and $\delta_i^{(2)}$ in cycle 1, and $\delta_i^{(2)}$ and $\delta_i^{(3)}$ in cycle 2. Thus the maximum number of right neuron parameters ever accessed in a cycle is $\left\lceil z_i/\indeg_i \right\rceil = 2$.

Since edges are interleaved on the left, in general, the $z_i$ edge processing logic units  will need access to $z_i$ parameters of a particular type from layer $i-1$. So all the left memory banks have $z_i$ memories, each of depth $D_i = N_{i-1}/z_i$, which are accessed in interleaved order. For example, after $D_i$ cycles, $N_{i-1}$ edges have been processed -- \ie $\left(D_i\times z_i\right) = N_{i-1}$.  We require that each of these edges be connected  to a different left neuron to eliminates the possibility of duplicate edges.   This completes a \emph{sweep}, \ie one complete access of the left memory bank. Since each left neuron connects to $\outdeg_i$ edges, $\outdeg_i$ sweeps are required to process all the edges, \ie each left activation is read $\outdeg_i$ times in the whole junction cycle. The reader can verify that $D_i$ cycles multiplied by $\outdeg_i$ sweeps results in $C_i$ total cycles, \ie one junction cycle.

\subsection{Clash-free connection patterns}\label{sec-hardware-clashfreedom}
We define a \emph{clash} as attempting to perform a particular operation more than once on the same memory at the same time, which would stall processing.\footnote{For single-ported memories, attempting two reads or two writes or a read and a write in the same cycle is a clash. For simple dual-ported memories with one port exclusively for reading and the other exclusively for writing, a read and a write can be performed in the same cycle. Attempting to perform two reads or two writes in the same cycle is a clash.} The idea of \emph{clash-freedom} is to pre-define a pattern of connections and $z$ values such that no operation in any junction of the \ac{NN} results in a clash. Sec. \ref{sec-hardware-memoryorg} described how $z$ values should be designed to prevent clashes in the weight and right memory banks.

This subsection analyzes the left memory banks, which are accessed in interleaved order. Their memory access pattern should be designed so as to prevent clashes. Additionally, the following properties are desired for practical clash-free patterns. Firstly, it should be easy to find a pattern that gives good performance. Secondly, the logic and storage required to generate the left memory addresses should be low complexity. 

We generate clash-free patterns by initially specifying the left memory addresses to be accessed in cycle 0 using a seed vector $\bm{\phi}_i \in \{0,1,\cdots,D_i-1\}^{z_i}$. Subsequent addresses are cyclically generated. Considering Fig. \ref{fig-memaccesses} as an example, $\bm{\phi}_i = (1,0,2,2)$. Thus in cycle 0, we access addresses $(1,0,2,2)$ from memories $(M0,M1,M2,M3)$, \ie left neurons $(4,1,10,11)$. In cycle 1, the accessed addresses are $(2,1,0,0)$, and so on. Since $D_i=3$, cycles 3--5 access the same left neurons as cycles 0--2.

We found that this technique results in a large number of possible connection patterns, as discussed in Appendix \ref{appendix-cftypes}. Randomly sampling from this set results in performance comparable with non-clash-free \ac{NN}s, as shown in Sec. \ref{sec-predefinedsparsity-comparison}. Finally, our approach only requires storing $\bm{\phi}_i$ and using $z_i$ incrementers to generate subsequent addresses. %power of 2 results in simpler hardware - Peter??
This approach is similar to methods used in modern coding to allow parallel processing and memory accesses, c.f. \cite{masera1999vlsi, crozier2001high, brack2007low}. 
Other techniques to generate clash-free connection patterns are discussed in Appendix \ref{appendix-cftypes}.

\subsection{Batch size}\label{sec-hardware-batchsize}
It is common in training of \ac{NN}s to use minibatches.  For a batch size of $M$, the \ac{UP} operation in \eqref{eq-up} is performed only once for $M$ inputs by using the average over the $M$ gradients.  Our architecture performs an \ac{UP} for every input and therefore may be viewed as having batch size one.  However, the processing in our architecture differs from a typical software implementation with $M=1$ due to the pipelined and parallel operations.  Specifically, in our architecture, \ac{FF} and \ac{BP} for the same input use different weights, as implied by Fig. \ref{fig-hardwareoverview}(c). In results not presented here, we found no performance degradation due to this variation from the standard backpropagation algorithm.  There is considerable ambiguity in the literature regarding ideal batch sizes \cite{Goyal2017,Masters2018}, and we found that our current network architecture performed well in our initial hardware implementation \cite{Dey2018_Reconfig}. However, if a more conventional batch size is desired, the \ac{UP} logic can be removed from the junction pipeline and the \ac{UP} operation performed once every $M$ inputs. This would eliminate some arithmetic units at the cost of increased storage for accumulating intermediate values from \eqref{eq-up}.
\begin{comment}
\begin{itemize}
    \item recent alternatives we're working on -- update in batch instead of always, more pipelining, ref Reconfig paper
    \item Analogy to high-speed FEC decoders
\end{itemize}
\end{comment}

\subsection{Special Case: Processing a \ac{FC} junction}
\begin{figure}[!t]
\centering
\includegraphics[width = 1.0\linewidth]{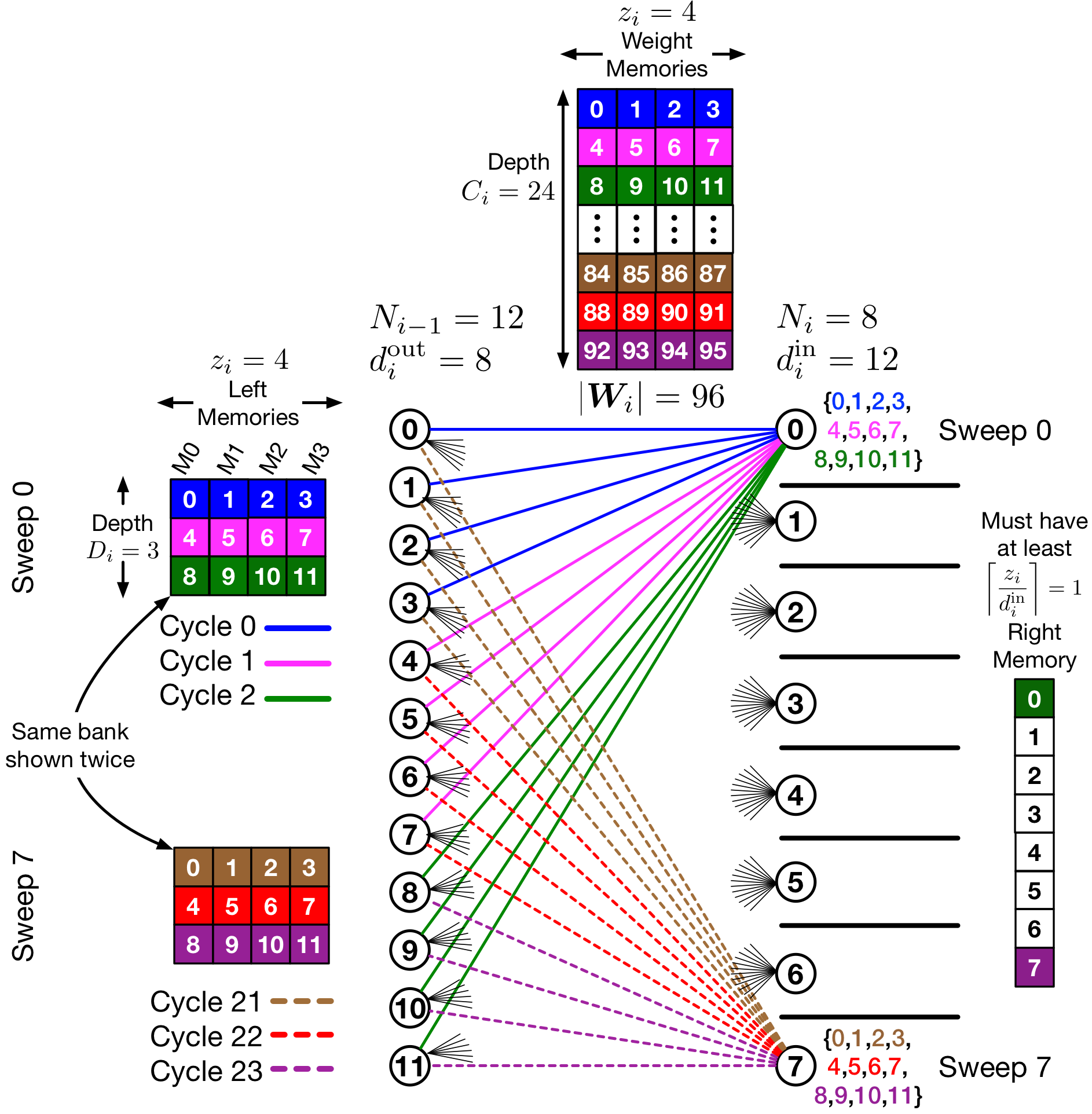}
\caption{Processing the \ac{FC} version of the junction from Fig. \ref{fig-memaccesses}. For clarity, only the first 12 and last 12 edges (dashed) are shown, corresponding respectively to right neurons 0 and 7, sweeps 0 and 7, cycles 0--2 and 21--23.}
\label{fig-memaccesses-fc}
\end{figure}

Fig. \ref{fig-memaccesses-fc} shows the \ac{FC} version of the junction from Fig. \ref{fig-memaccesses}, which has 96 edges to be accessed and operated on. This can be done keeping the same junction cycle $C_i=6$ by increasing $z_i$ to 16, \ie using more hardware. On the other hand, if hardware resources are limited, one can use the same $z_i=4$ and pay the price of a longer junction cycle $C_i=24$, as shown in Fig. \ref{fig-memaccesses-fc}. This demonstrates the flexibility of our architecture.

Note that \ac{FC} junctions are clash-free in all practical cases due to the following reasons. Firstly, the left memory accesses are in natural order just like the weights, which ensures that no more than one element is accessed from each memory per cycle. Secondly, $\left\lceil z_i/\indeg_i \right\rceil = 1$ for all practical cases since $z_i\leq N_{i-1}$, as discussed in Appendix \ref{appendix-hardware-constraints}, and $\indeg_i=N_{i-1}$ for \ac{FC} junctions. This means that at most one right neuron is processed in a cycle\footnote{In Fig. \ref{fig-memaccesses-fc} for example, one right neuron finishes processing every $3^{\mathrm{rd}}$ cycle}, so clashes will never occur when accessing the right memory bank.

Note that compared to Fig. \ref{fig-memaccesses}, the weight memories in Fig. \ref{fig-memaccesses-fc} are deeper since $C_i$ has increased from 6 to 24. However, the left layer memories remain the same size since $N_{i-1}=12$ and $z_i=4$ are unchanged, but the left memory bank is accessed more times since the number of sweeps has increased from 2 to 8. Also note that even if cycle 0 (blue) accesses some other clash-free subset of left neurons, such as $\{4,5,6,7\}$ instead of $\{0,1,2,3\}$, the connection pattern would remain unchanged. This implies that different memory access patterns do not necessarily lead to different connection patterns; as discussed further in Appendix \ref{appendix-cftypes}.

\section{Observed Trends of Pre-Defined Sparsity}\label{sec-trends}
This section analyzes trends  observed when experimenting with several different datasets via software simulations. We intend the following four trends to provide guidelines on designing pre-defined sparse \ac{NN}s.
\begin{enumerate}
\item Hardware-compatible, clash-free, pre-defined sparse patterns perform at least as well as other pre-defined sparse patterns (\ie random and structured) (Sec. \ref{sec-predefinedsparsity-comparison}).
\item The performance of pre-defined sparsity is better on datasets that have more inherent redundancy (Sec. \ref{sec-trends-redundancy}).
\item  Junction density should increase to the right: junctions closer to the output should generally have more connections than junctions closer to the input (Sec. \ref{sec-trends-jndensities}).  
\item Larger and more sparse \ac{NN}s are better than smaller and denser \ac{NN}s, given the same number of layers and trainable parameters. Specifically,  `larger' refers to more hidden neurons (Sec. \ref{sec-trends-lsvsd}). 
\end{enumerate}
The remainder of this section first describes the datasets we experimented on, and then examines these trends in detail.

\subsection{Datasets and Experimental Configuration}\label{sec-datasets}
Unless otherwise noted, the following parameters and configurations listed below were used for all presented results.  
\paragraph{MNIST handwritten digits} We rasterized each input image into a single layer of 784 features\footnote{On certain occasions we added 16 input features which are always trivially 0 so as to get 800 features for each input. This leads to easier selection of different sparse network configurations.}, \ie the permutation-invariant format. No data augmentation was applied.
\paragraph{Reuters RCV1 corpus of newswire articles} The classification categories are grouped in a tree structure. We used  preprocessing techniques similar to \cite{Hinton2012} to isolate articles which fell under a single category at the second level of the tree. We finally obtained 328,669 articles in 50 categories, split into $50,000$ for validation, $100,000$ for test, and the remaining for training. The original data has a list of token strings for each story, for example, a story on finance would frequently contain the token `financ'. We chose the most common 2000 tokens and computed counts for each of these in each article. Each count $x$ was transformed into $\text{log}(1+x)$ to form the final 2000-dimensional feature vector for each input.
\paragraph{TIMIT speech corpus} TIMIT is a speech dataset comprising approximately $5.4$ hours of 16 kHz audio commonly used in \ac{ASR}.  A modern \ac{ASR} system has three major components: (i) preprocessing and feature extraction, (ii) acoustic model, and (iii) dictionary and language model.  A complete study of an ASR system is beyond the scope of this work.  Instead we focus on the acoustic model which is typically implemented using a \ac{NN}.  The input to the acoustic model is feature vectors and the output is a probability distribution on phonemes (\ie speech sounds).  %The sequence of phoneme distributions are mapped to sentences using the dictionary and language model.
For our experiments, we used 25ms speech frames with 10ms shift, as in \cite{Hinton2012}, and computed a feature vector of 39 \ac{MFCC}s for each frame. We used the complete training set of $818,837$ training samples (462 speakers), $89,319$ validation samples (50 speakers), and $212,093$ test samples (118 speakers). We used a phoneme set of size 39 as defined in \cite{Lee1989_TIMIT}.  

\paragraph{CIFAR-100 images} Our setup for CIFAR-100 consists of a \ac{CNN} followed by a \ac{MLP}. The \ac{CNN} has 3 blocks and each block has 2 convolutional layers with window size 3x3 followed by a max pooling layer of pool size 2x2. The number of filters for the six convolutional layers is (60,60, 125,125, 250,250). This results in a total of approximately one  million trainable parameters in the convolutional portion of the network. Batch normalization is applied before activations. The output from the 3rd block, after flattening into a vector, has 4000 features. Typically dropout is applied in the \ac{MLP} portion, however we omitted it there since pre-defined sparsity is an alternate form of parameter reduction. Instead we found that a dropout probability of half applied to the convolutional blocks improved performance. No data augmentation was applied.

\begin{comment}
    In other words,
    \begin{IEEEeqnarray}{c}
    \text{TPC} = \frac{1}{212093}\sum_{i=1}^{212093}{\sum_{j=1}^{39}{A_{ij}\text{log}\left(\frac{A_{ij}}{B_{ij}}\right)}}
    \end{IEEEeqnarray}
    where $\underset{212093\times39}{A}$ and $\underset{212093\times39}{B}$ are the complete output test matrices for the $212,093$ test samples with 39 labels each for the \ac{FC} and sparse cases, respectively. Lesser values of \ac{TPC} are better as they indicate minimal performance degradation due to sparsification.
\end{comment}

For each dataset, we performed classification using one-hot labels and measured accuracy on the test set as a performance metric.\footnote{The \ac{NN} in a complete \ac{ASR} system would be a `soft' classifier and feed the phoneme distribution outputs to a decoder to perform `hard' final classification decisions. Therefore for TIMIT, we computed another performance metric called \ac{TPC}, measured as KL divergence between predicted test output probability distributions of sparse vs the respective \ac{FC} case. Performance results obtained using \ac{TPC} were qualitatively very similar to test accuracy and not shown here.}  We also calculated the top-5 test set classification accuracy for CIFAR-100.

We found the optimal training configuration for each \ac{FC} setup by doing a grid search using validation performance as a metric. This resulted in choosing ReLU activations for all layers except for the final softmax layer. The initialization proposed by He \textit{et al.} \cite{He2015_wtinit} worked best for the weights; while for biases, we found that an initial value of $0.1$ worked best in all cases except for Reuters, for which zeroes worked better.  The Adam optimizer \cite{Kingma2014_Adam} was used with all parameters set to default, except that we set the decay parameter to $10^{-5}$ for best results. We used a batch size of 1024 for TIMIT and Reuters since the number of training samples is large, and 256 for MNIST and CIFAR.

All experiments were run for 50 epochs of training and regularization was applied as an L2 penalty to the weights. To maintain consistency, we kept most hyperparameters the same when sparsifying the network, but reduced the L2 penalty coefficient with increasing sparsity. This was done because sparse \ac{NN}s have  fewer trainable parameters and are less prone to overfitting.  We ran each experiment  at least five times to average out randomness and we show the 90\% \ac{CI}s for each metric as shaded regions (this also holds for the results in Fig. \ref{fig-pdsexample}(c,h)).  In addition to the results shown, we developed a data set of Morse code symbol sequences and investigated pre-defined sparse \ac{NN}s.   While these results are excluded for brevity, they are consistent with the trends described in this Section, and can be found in \cite{Dey2018_ICCCNT}.

\subsection{Comparison of Pre-Defined Sparse Methods}\label{sec-predefinedsparsity-comparison}

\begin{table}[!t]
\renewcommand{\arraystretch}{1.1}
\caption{Comparison of Pre-Defined Sparse Methods}
\label{table-predefinedsparsity-comparison}
\centering
\begin{minipage}{\columnwidth}
\resizebox{\columnwidth}{!}{
\begin{tabular}{|c|c|c|c|c|c|}
\hline
\multirow{2}{*}{$\outdegnet$} & \multirow{2}{*}{$\rhonet \%$} & \multirow{2}{*}{$\znet$} & \multicolumn{3}{|c|}{Test Accuracy Performance} \\
\cline{4-6}
& & & Clash-free & Structured & Random \\
\hline
\hline
\multicolumn{6}{|c|}{\multirow{2}{*}{MNIST: $\Nnet = (800,100,100,100,10)$, \ac{FC} test accuracy = $98\pm0.1$}} \\
\multicolumn{6}{|c|}{} \\
\hline
$(80,80,80,10)$ & $80.2$ & $(200,25,25,4)$ & $97.9\pm0.2$ & $97.9\pm0.2$ & $97.8\pm0.2$ \\
\hline
$(60,60,60,10)$ & $60.4$ & $(200,25,25,4)$ & $97.6\pm0.1$ & $97.8\pm0.1$ & $97.6\pm0.2$ \\
\hline
$(40,40,40,10)$ & $40.6$ & $(200,25,25,5)$ & $97.5\pm0.1$ & $97.7$ & $97.6\pm0.1$ \\
\hline
$(20,20,20,10)$ & $20.8$ & $(200,25,25,10)$ & $97.2\pm0.2$ & $97.2\pm0.1$ & $97.1\pm0.1$ \\
\hline
$(10,10,10,10)$ & $10.9$ & $(200,25,25,25)$ & $96.7\pm0.1$ & $96.8\pm0.2$ & $96.7\pm0.2$ \\
\hline
$(5,10,10,10)$ & $6.9$ & $(100,25,25,25)$ & $96.3\pm0.1$ & $96.3\pm0.1$ & $96.2\pm0.1$ \\
\hline
$(2,5,5,10)$ & $3.6$ & $(80,25,25,50)$ & $95\pm0.2$ & $95.1\pm0.1$ & $95\pm0.3$ \\
\hline
$(1,2,2,10)$ & \textcolor{blue}{$2.2$} & $(80,20,20,100)$ & \textcolor{blue}{$93.3\pm0.3$} & \textcolor{blue}{$93.1\pm0.5$} & \textcolor{blue}{$92\pm0.3$} \\
\hline
\multicolumn{6}{|c|}{\multirow{2}{*}{Reuters: $\Nnet = (2000,50,50)$, \ac{FC} test accuracy = $89.6\pm0.1$}} \\
\multicolumn{6}{|c|}{} \\
\hline
$(25,25)$ & $50$ & $(1000,25)$ & $89.4\pm0.1$ & $89.3$ & $89.4$ \\
\hline
$(10,10)$ & $20$ & $(400,10)$ & $87\pm0.1$ & $86.7\pm0.1$ & $86.5\pm0.1$ \\
\hline
$(5,5)$ & $10$ & $(200,5)$ & $78.5\pm0.5$ & $78.2\pm0.7$ & $77.5\pm0.6$ \\
\hline
$(2,2)$ & \textcolor{blue}{$4$} & $(80,2)$ & \textcolor{blue}{$53.3\pm1.8$} & \textcolor{blue}{$51.2\pm1.7$} & \textcolor{blue}{$46.8\pm2.9$} \\
\hline
$(1,1)$ & $2$ & $(40,1)$ & $28.4\pm2.4$ & $28.7\pm2.3$ & $28\pm1.9$ \\
\hline
\multicolumn{6}{|c|}{\multirow{2}{*}{TIMIT: $\Nnet = (39,390,39)$, \ac{FC} test accuracy = $43.2\pm0.2$}} \\
\multicolumn{6}{|c|}{} \\
\hline
$(270,27)$ & $69.2$ & \multirow{5}{*}{$(13,13)$} & $43\pm0.1$ & $43$ & $43\pm0.1$ \\
\cline{1-2}\cline{4-6}
$(180,18)$ & $46.2$ & & $42.7\pm0.1$ & $42.8\pm0.1$ & $42.9\pm0.1$ \\
\cline{1-2}\cline{4-6}
$(90,9)$ & $23.1$ & & $42.1\pm0.1$ & $42.5\pm0.1$ & $42.4\pm0.1$ \\
\cline{1-2}\cline{4-6}
$(60,6)$ & $15.4$ & & $41.5\pm0.1$ & $41.8\pm0.2$ & $41.9\pm0.1$ \\
\cline{1-2}\cline{4-6}
$(30,3)$ & \textcolor{blue}{$7.7$} & & \textcolor{blue}{$40.5\pm0.2$} & \textcolor{blue}{$40.1\pm0.2$} & \textcolor{blue}{$39.4\pm0.8$} \\
\hline
\multicolumn{6}{|c|}{\multirow{2}{*}{CIFAR-100 \footnote{For CIFAR-100, given values of $\Nnet$, $\outdegnet$, $\znet$ and $\rhonet$ are just for the \ac{MLP} portion, which follows a \ac{CNN} as described in Sec. \ref{sec-datasets} to form the complete net. Reported values are top-5 test accuracies obtained from training on the complete net.}: $\Nnet = (4000,500,100)$, \ac{FC} top-5 test accuracy = $87.1\pm0.6$}} \\
\multicolumn{6}{|c|}{} \\
\hline
$(100,100)$ & $22$ & \multirow{2}{*}{$(2000,250)$} & $87.5\pm0.2$ & $87.7\pm0.2$ & $87.4\pm0.3$ \\
\cline{1-2}\cline{4-6}
$(29,29)$ & $6.4$ & & $86.8\pm0.3$ & $87.2\pm0.5$ & $87.1\pm0.2$ \\
\hline
$(12,12)$ & $2.6$ & \multirow{2}{*}{$(400,50)$} & $86.3\pm0.2$ & $86.5\pm0.4$ & $86.6\pm0.4$ \\
\cline{1-2}\cline{4-6}
$(5,5)$ & $1.1$ & & $85.3\pm0.5$ & $85.5\pm0.5$ & $85.7\pm0.3$ \\
\hline
$(2,2)$ & $0.4$ & \multirow{2}{*}{$(80,10)$} & $84.1\pm0.5$ & $84.3\pm0.3$ & $83.8\pm0.3$ \\
\cline{1-2}\cline{4-6}
$(1,1)$ & \textcolor{blue}{$0.2$} & & \textcolor{blue}{$83\pm0.5$} & \textcolor{blue}{$83.3\pm0.4$} & \textcolor{blue}{$81.7\pm0.7$} \\
\hline
\end{tabular}
}
\end{minipage}
\end{table}

Table \ref{table-predefinedsparsity-comparison} shows performance on different datasets for three methods of pre-defined sparsity: a) the most restrictive and hardware-friendly clash-freedom, b) structured, and c) random. For the clash-free case, we experimented with different $\znet$ settings to simulate different hardware environments:%\footnote{$\znet$ is always chosen to obey the constraint of same junction cycle length for all junctions. The only exceptions are the 1st 4 MNIST networks where the numbers are slightly mismatched.}
\begin{itemize}
    \item Reuters: One junction cycle is 50 cycles for all the different densities. This is because we scale $\znet$ accordingly, \ie a more powerful hardware device is used for each \ac{NN} as $\rhonet$ increases.
    \item CIFAR-100 and MNIST: These simulate cases where hardware choice is limited, such as a high-end, a mid-range and a low-end device being available. Thus three different $\znet$ values are used for CIFAR-100 depending on $\rhonet$.
    \item TIMIT: We keep $\znet$ constant for different densities. Junction cycle length varies from 90 cycles for $\rhonet = 7.69\%$ to 810 for $\rhonet = 69.23\%$. This shows that when limited to a single low-end hardware device, denser \ac{NN}s can be processed in longer time by simply changing $\znet$.
\end{itemize}

Table \ref{table-predefinedsparsity-comparison} confirms that \emph{hardware-friendly clash-free pre-defined sparse architectures do not lead to any statistically significant performance degradation}. We also observed  that random pre-defined sparsity performs poorly for very low density networks, as shown by the blue values. This is possibly because there is non-negligible probability of neurons getting completely disconnected, leading to irrecoverable loss of information. %Also note the large $90\%$ \ac{CI} for the random cases for these points. This is possibly because the set of disconnected neurons vary widely in importance. In MNIST for example, a pixel from the center of an image getting disconnected will lead to worse results than a corner pixel getting disconnected (this leads to attention based sparsity described in Sec. \ref{sec-othersparsity-attention}). This observation helps to justify our reasons in Sec. \ref{sec-spds-basics} for choosing fixed $\indeg$ and $\outdeg$.

\subsection{Dataset Redundancy}\label{sec-trends-redundancy}
Many machine learning datasets have considerable redundancy in their input features. For example, one may not need information from the $\sim$800 input features of MNIST to infer the correct image class. We hypothesize that pre-defined sparsity takes advantage of this redundancy, and will be less effective when the redundancy is reduced. To test this, we changed the feature vector for each dataset as follows. For MNIST, \ac{PCA} was used to reduce the feature count to the least redundant 200. For Reuters, the number of most frequent tokens considered as features was reduced from 2000 to 400. For TIMIT, we both reduced and increased the number of \ac{MFCC}s by 3X to 13 and 117, respectively. Note that the latter increases redundancy. For CIFAR-100, a source of redundancy is the depth of the \ac{CNN}, which extracts features and discriminates between classes before the \ac{MLP} performs final classification. In other words, the \ac{CNN} eases the burden of the \ac{MLP}. So a way to reduce redundancy and increase the classification burden of the \ac{MLP} is to lessen the effectiveness of the \ac{CNN} by reducing its depth. Accordingly, we used a single convolutional layer with 250 filters of window size $5\times5$ followed by a $8\times8$ max pooling layer. This results in the same number of features, 4000, at the input of the \ac{MLP} as the original network, but has reduced redundancy for the \ac{MLP}.

\begin{figure}[!t]
\centering
\includegraphics[width = 1.0\linewidth]{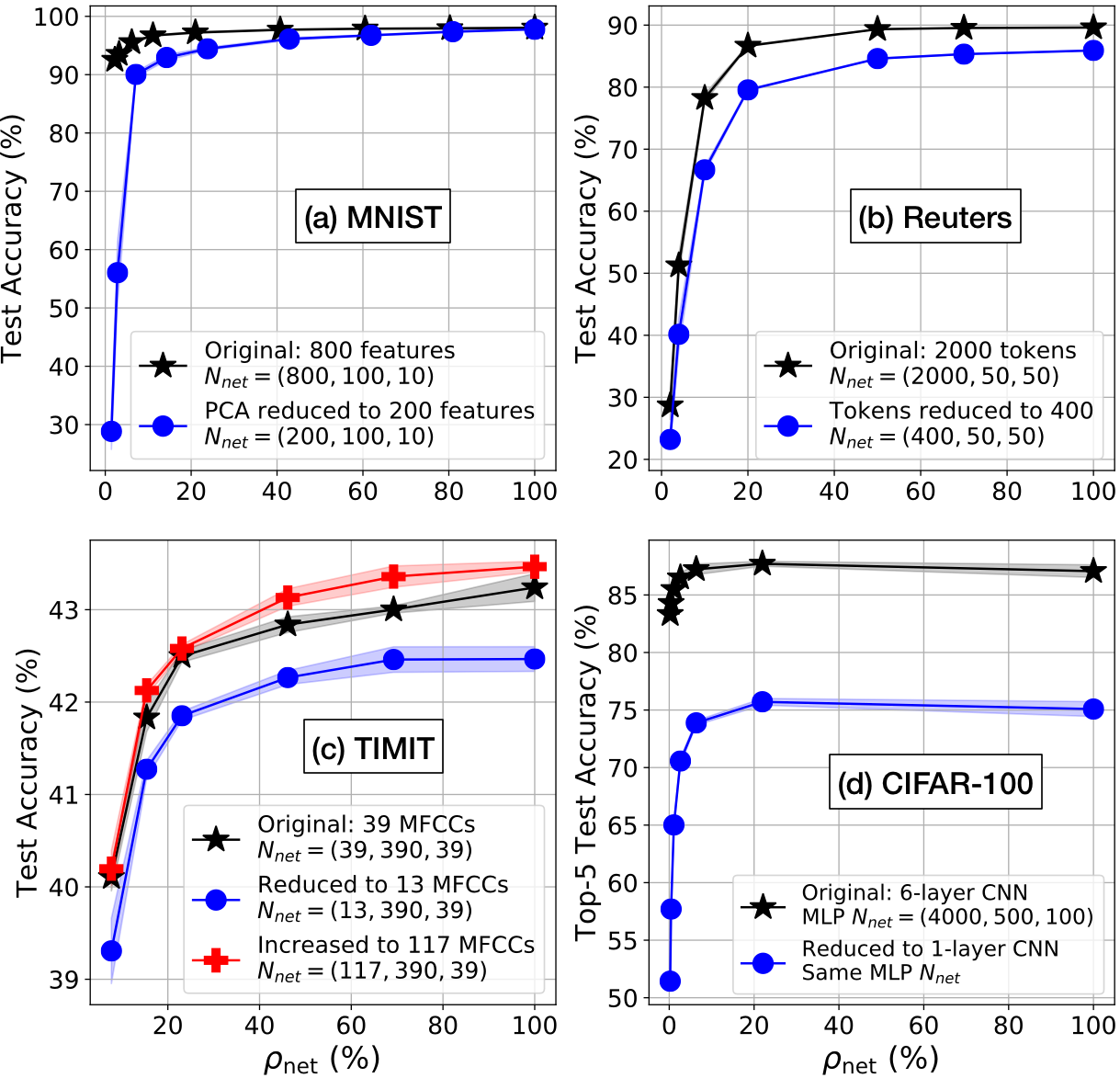}
\caption{Comparison of classification accuracy as a function of  $\rhonet$ for different versions of datasets -- original, reduced in redundancy by reducing feature space (MNIST, Reuters, TIMIT) or performing less processing prior to the \ac{MLP} (CIFAR-100), and increasing redundancy by enlarging feature space (TIMIT).}
\label{fig-originalvsreduced_sp}
\end{figure}

Classification performance results are shown in Fig. \ref{fig-originalvsreduced_sp} as a function of $\rhonet$.  For MNIST and CIFAR-100, the performance degrades more sharply with reducing $\rhonet$ for the nets using the reduced redundancy datasets. To explore this further, we recreated the histograms from Fig. \ref{fig-pdsexample} for the reduced redundancy datasets, \ie a \ac{FC} \ac{NN} with $\Nnet=(200,100,10)$ training on MNIST after \ac{PCA}. We observed a wider spread of weight values, implying less opportunity for sparsification (\ie fewer weights were close to zero). Similar trends are less discernible for Reuters and TIMIT, however, reducing redundancy led to worse performance overall.  

The results in Fig. \ref{fig-originalvsreduced_sp} further demonstrate the effectiveness of pre-defined sparsity in greatly reducing network complexity with negligible performance degradation. For example, even the reduced redundancy problems perform well when operating with half the number of connections. For CIFAR in particular, \emph{\ac{FC} performs worse than an overall \ac{MLP} density of around 20\%}. Thus, in addition to reducing complexity, structured pre-defined sparsity may be viewed as an alternative to dropout in the \ac{MLP} for the purpose of improving classification performance.

\subsection{Individual junction densities}\label{sec-trends-jndensities}

\begin{figure}[!t]
\centering
\includegraphics[width = 1.0\linewidth]{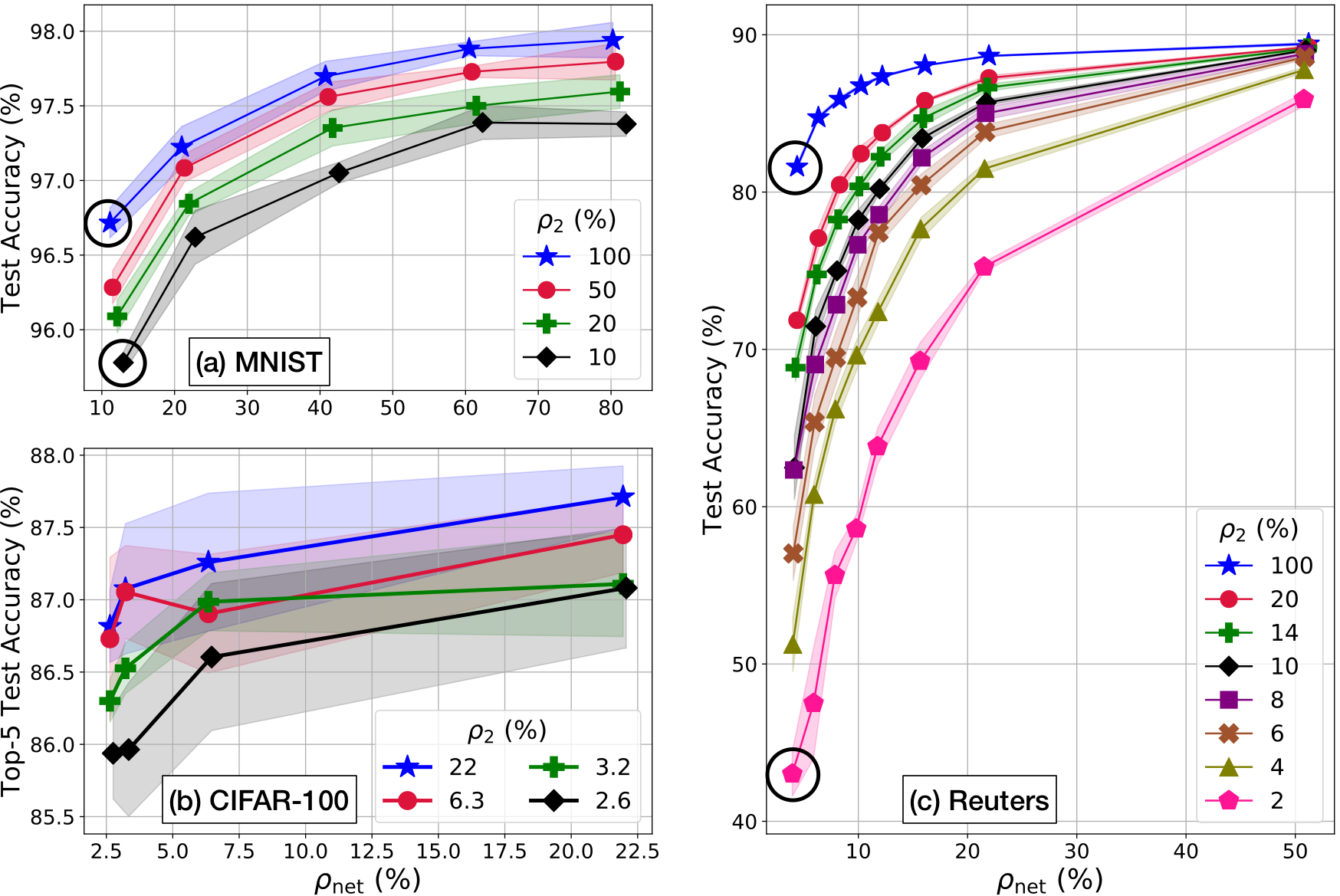}
\caption{Comparison of classification accuracy as a function of   $\rhonet$ for different $\rho_L$, where $L=2$. Black-circled points show the effects of $\rho_2$ when $\rhonet$ is the same. $\Nnet$ values are $(800,100,10)$ for MNIST, $(2000,50,50)$ for Reuters, and $(4000,500,100)$ for the \ac{MLP} in CIFAR-100.}
\label{fig-jnimp_mnist_rcv1_cifar}
\end{figure}

The weight histograms in Fig. \ref{fig-pdsexample} indicate that latter junctions, particularly junction $L$ closest to the output, have a wide spread of weight values. This suggests that a good strategy for reducing $\rhonet$ would be to use lower densities in earlier junctions -- \ie $\rho_1<\rho_L$. This is demonstrated in Fig. \ref{fig-jnimp_mnist_rcv1_cifar} for the cases of MNIST, CIFAR-100 and Reuters, each with $L=2$ junctions in their \ac{MLP}s. Each curve in each subfigure is for a fixed $\rho_2$, \ie reducing $\rhonet$ across a curve is done solely by reducing $\rho_1$. For a fixed $\rhonet$, the performance improves as $\rho_2$ increases. For example, the circled points in Reuters both have $\rhonet=4\%$, but the starred point with $\rho_2=100\%$ has approximately  $40\%$ better test accuracy than the pentagonal point with $\rho_2=2\%$. The trend  clearly holds for MNIST and is also discernible for CIFAR-100.

\begin{figure}[!t]
\centering
\includegraphics[width = 1.0\linewidth]{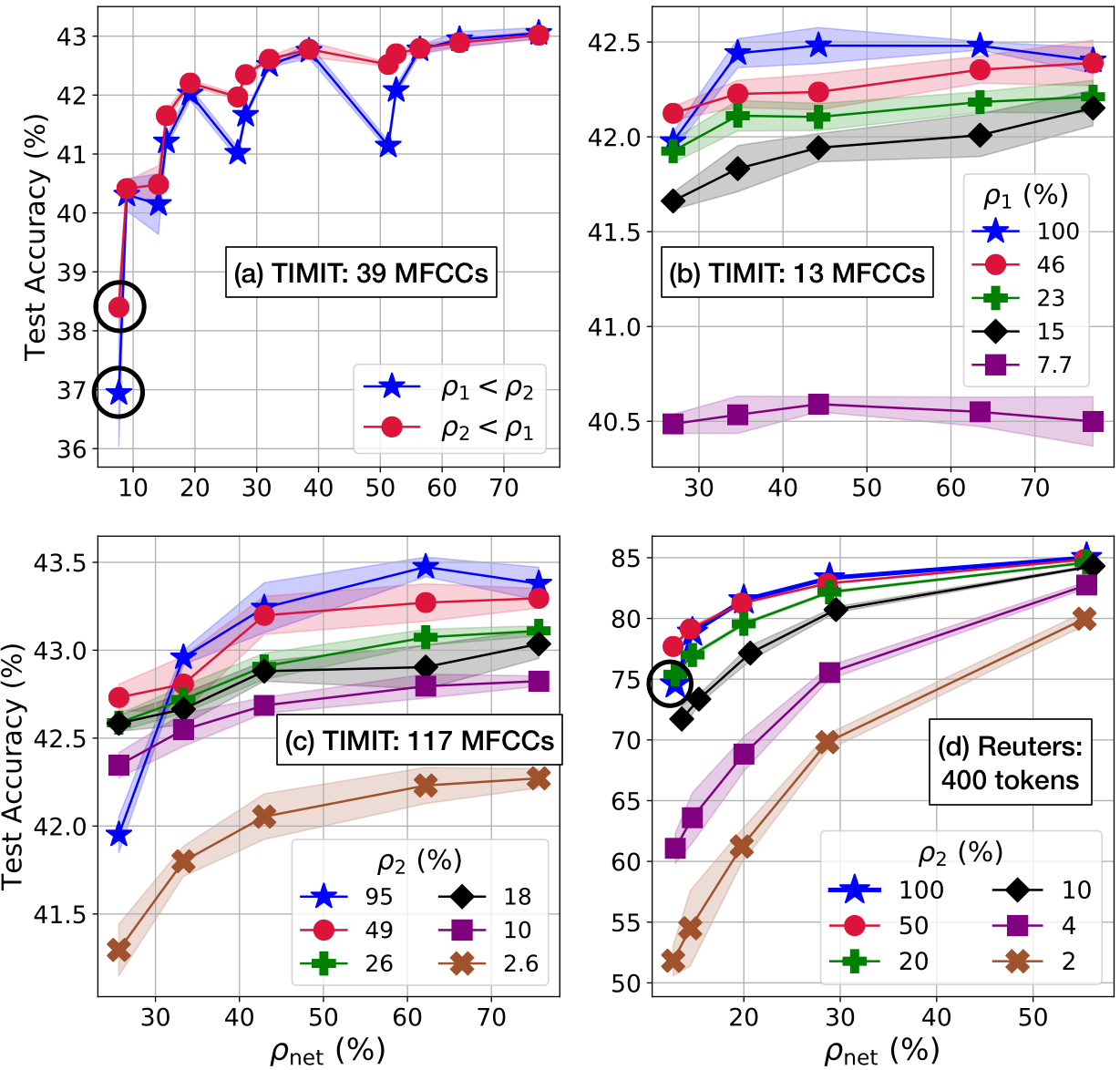}
\caption{Comparison of  classification accuracy as a function of  $\rhonet$ for: (a) TIMIT with 39 \ac{MFCC}s for the two cases where one junction is always sparser than the other and vice-versa. Black-circled points show how reducing $\rho_1$ degrades performance to a greater extent. (b) TIMIT with 13 \ac{MFCC}s for different $\rho_1$. (c,d) TIMIT with 117 \ac{MFCC}s, and Reuters reduced to 400 tokens, for different $\rho_2$. $\Nnet$ values are (a) $(39,390,39)$, (b) $(13,390,39)$, (c) $(117,390,39)$, (d) $(400,50,50)$.}
\label{fig-jnimp_timit_rrcv1}
\end{figure}

We further observed that this trend (\ie $\rho_{i+1} > \rho_i$ should hold) is related to the redundancy inherent in the dataset and may not hold for datasets with very low levels of redundancy.  To explore this, results analogous to those in Fig. \ref{fig-jnimp_mnist_rcv1_cifar} are presented in Fig. \ref{fig-jnimp_timit_rrcv1} for TIMIT,  but with varying sized \ac{MFCC} feature vectors -- \ie datasets corresponding to larger feature vectors will contain more redundancy.  The results in Fig. \ref{fig-jnimp_timit_rrcv1}(c) are for 117 dimensional \ac{MFCC}s and are consistent with the trend in  Fig. \ref{fig-jnimp_mnist_rcv1_cifar}.  However, for a \ac{MFCC} dimension of 13, this trend actually reverses -- \ie the junction 1 should have higher density. This is shown in Fig. \ref{fig-jnimp_timit_rrcv1}(b), where each curve is for a fixed $\rho_1$.  This reversed trend is also observed for the case of 39 dimensional feature vectors,  considered in Fig. \ref{fig-jnimp_timit_rrcv1}(a), where $\Nnet = (39,390,39)$.  Due to this symmetric neuronal configuration, for each value of $\rhonet$ on the x-axis in Fig. \ref{fig-jnimp_timit_rrcv1}(a), the two curves have complementary values of $\rho_1$ and $\rho_2$ ($\rho_1\ne\rho_2$) -- \eg the two curves at $\rhonet = 7.69\%$ have  $(\rho_1, \rho_2)$ pairs of $(2.56\%, 12.82\%)$ and $(12.82\%, 2.56\%)$. We observe that the curve for $\rho_1<\rho_2$ is generally worse than the curve for $\rho_2<\rho_1$, which indicates that junction 1 should have higher density in this case.

Fig. \ref{fig-jnimp_timit_rrcv1}(d) depicts the results for Reuters with the feature vector size reduced to 400 tokens. While junction 2 is still more important (as in Fig. \ref{fig-jnimp_mnist_rcv1_cifar}(c) for the original Reuters dataset), notice the circled star-point at the very left of the $\rho_2=100\%$ curve. This point has very low $\rho_1$. Unlike Fig. \ref{fig-jnimp_mnist_rcv1_cifar}(c), it crosses below the other curves, indicating that it is more important to have higher density in the first junction with this less redundant set of features.  
We observed a similar, but less prominent, trend in MNIST \ac{PCA} when the feature dimension was reduced  to 200.

In summary, if an individual junction density falls below a certain value, referred to as the \emph{critical junction density}, it will adversely affect performance regardless of the density of other junctions. This explains why some of the curves cross in Fig. \ref{fig-jnimp_timit_rrcv1}. The critical junction density is much smaller for earlier junctions than for later junctions in most datasets with sufficient redundancy. However, the critical density for earlier junctions increases for datasets with low redundancy.

\subsection{`Large and sparse' vs `small and dense' networks}\label{sec-trends-lsvsd}
We observed that when keeping the total number of trainable parameters the same, sparser \ac{NN}s with larger hidden layers (\ie more neurons) generally performed better than denser networks with smaller hidden layers. This is true as long as the larger \ac{NN} is not so sparse that individual junction densities fall below the critical density, as explained in Sec. \ref{sec-trends-jndensities}. While the critical density is problem-dependent, it is usually low enough to obtain significant complexity savings above it. Thus, `large and sparse' is better than `small and dense' for many practical cases, including \ac{NN}s with more than one hidden layer (\ie $L>2$).

\begin{figure}[!t]
\centering
\includegraphics[width = 1.0\linewidth]{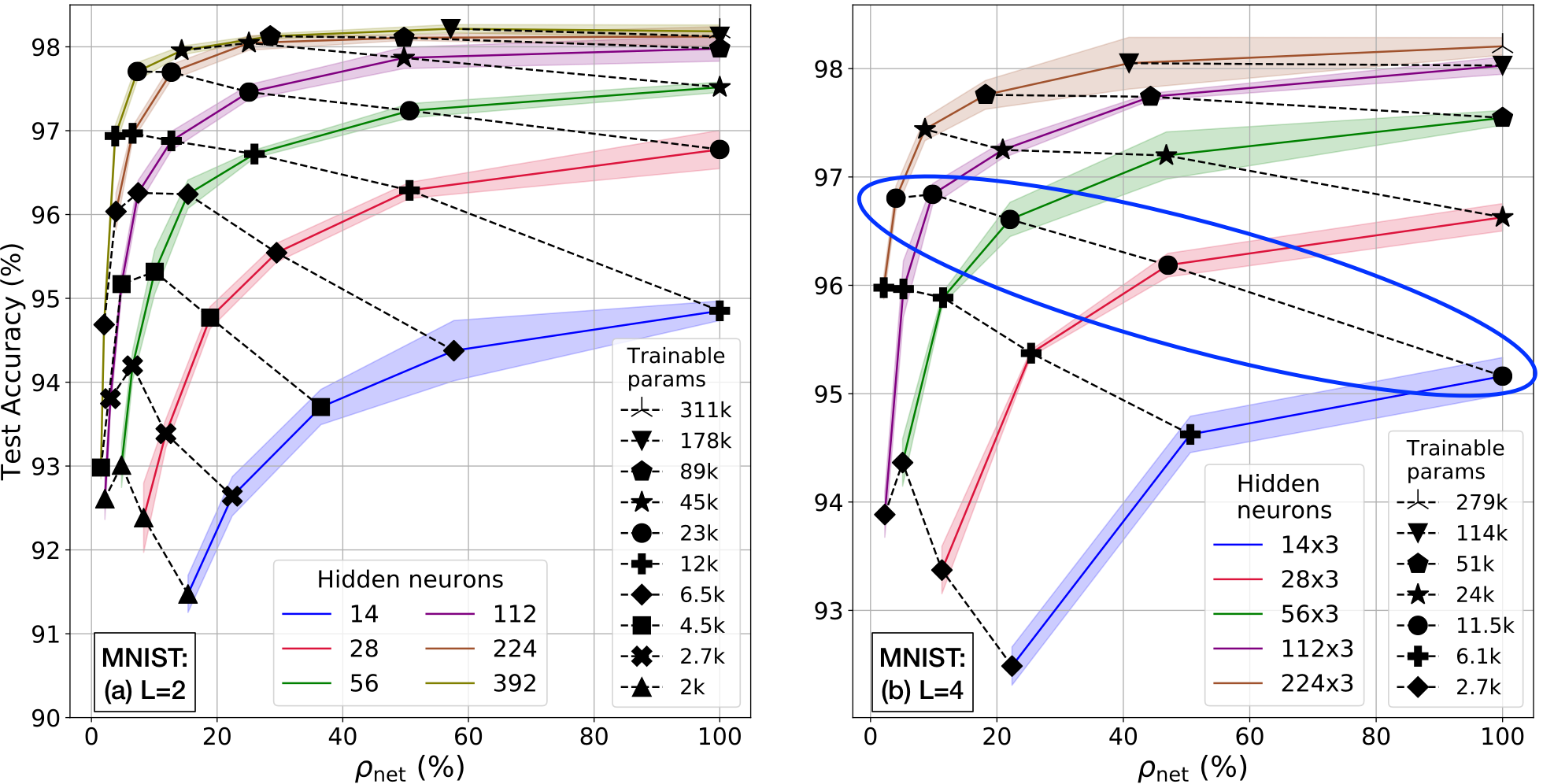}
\caption{Comparing `large and sparse' to `small and dense' networks for MNIST with 784 features, with (a) $\Nnet=(784,x,10)$ (on the left), and (b) $\Nnet=(784,x,x,x,10)$ (on the right). Solid curves (with the shaded \ac{CI}s around them) are for constant $x$, black dashed curves with same marker are for same number of trainable parameters. The final junction is always \ac{FC}. Intermediate junctions for the $L=4$ case have $\outdeg$ values similar to junction 1.}
\label{fig-lsvsd_mnist}
\end{figure}

Fig. \ref{fig-lsvsd_mnist} shows this for networks having one and three hidden layers trained on MNIST. For the three layer network, all hidden layers have the same number of neurons. Each solid curve shows classification performance vs $\rhonet$ for a particular $\Nnet$, while the black dashed curves with identical markers are configurations that have approximately the same number of trainable parameters. As an example,  the points with circular markers (with a big blue ellipse around them) in Fig. \ref{fig-lsvsd_mnist}(b)
all have the same number of trainable parameters and indicate that the larger, more sparse \ac{NN}s perform better.  Specifically, the network with $\Nnet=(784,112,112,112,10)$ and $\outdegnet=(10,10,10,10)$ corresponding to $\rhonet$ $=9.82\%$ performs significantly better than the \ac{FC} network with $\Nnet=(784,14,14,14,10)$, and other smaller and denser networks, despite each having $11500$ trainable parameters. Increasing the network size further to $\Nnet=(784,224,224,224,10)$, and reducing $\rhonet$ to $4\%$ to fix the number of trainable parameters at $11500$,  leads to performance degradation.   This is because this $\rhonet$ was achieved by setting $\rho_2=\rho_3=2.68\%$, which appears to be below the critical density.

\begin{figure}[!t]
\centering
\includegraphics[width = 1.0\linewidth]{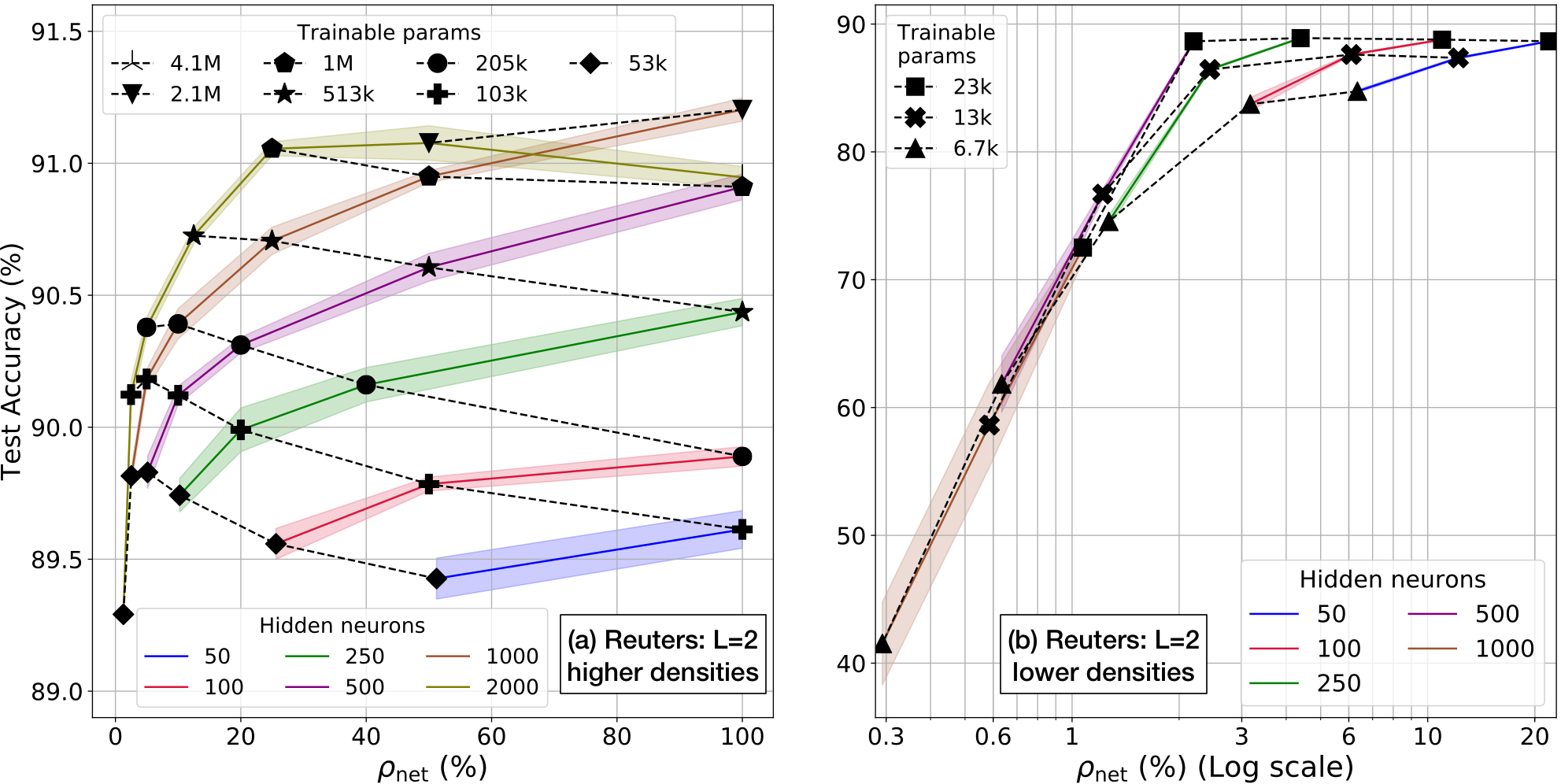}
\caption{Comparing `large and sparse' to `small and dense' networks for Reuters with 2000 tokens, with $\Nnet=(2000,x,50)$. The x-axis is split into higher values on the left (a), and lower values on the right in log scale (b). Solid curves (with the shaded \ac{CI}s around them) are for constant $x$, black dashed curves with same marker are for same number of trainable parameters. Junction 1 is sparsified first until its number of total weights is approximately equal to that of junction 2, then both are sparsified equally.}
\label{fig-lsvsd_rcv1}
\end{figure}

Fig. \ref{fig-lsvsd_rcv1} summarizes the analogous  experiment on Reuters with similar conclusions. Both subfigures are for the same results with the x-axis split into higher and lower density range (on log scale), to show more detail. Observe that the trend of `large and sparse' being better than `small and dense' holds for subfigure (a), but reverses for (b) since densities are very low (the black dashed curves have positive slope instead of negative). This is due to the critical density effect.

\begin{figure}[!t]
\centering
\includegraphics[width = 1.0\linewidth]{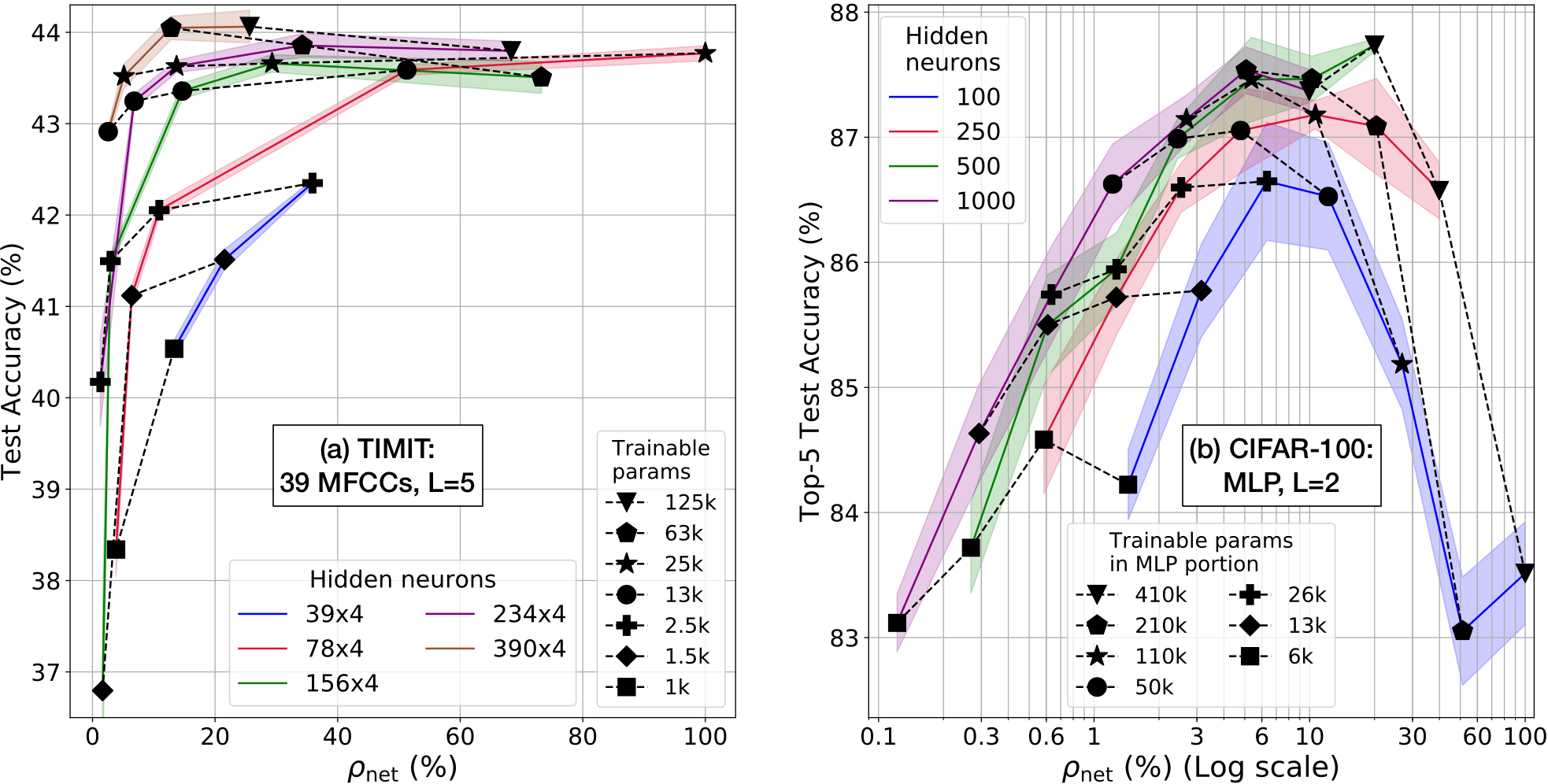}
\caption{Comparing `large and sparse' to `small and dense' networks for (a) TIMIT with 39 \ac{MFCC}s and $\Nnet=(39,x,x,x,x,39)$ (on the left), and (b) CIFAR-100 with the deep 6-layer \ac{CNN} and \ac{MLP} $\Nnet = (4000,x,100)$ with log scale for the x-axis (on the right). Solid curves (with the shaded \ac{CI}s around them) are for constant $x$, black dashed curves with same marker are for same number of trainable parameters (in the \ac{MLP} portion only for CIFAR). Since TIMIT has symmetric junctions, we tried to keep input and output junction densities as close as possible and adjusted intermediate junction densities to get the desired $\rhonet$. CIFAR-100 is sparsified in a way similar to Reuters in Fig. \ref{fig-lsvsd_rcv1}.}
\label{fig-lsvsd_timit_cifar}
\end{figure}

Fig. \ref{fig-lsvsd_timit_cifar}(a) shows the result for the same experiment  on TIMIT   with four hidden layers\footnote{We also performed experiments on TIMIT with one hidden layer ($L=2$) and Reuters with 2 hidden layers ($L=3$). Results were similar to those shown, so are not shown for brevity's sake.}. The trend is less clearly discernible, but it exists. Notice how the black dashed curves have negative slopes at appreciable levels of $\rhonet$, indicating `large and sparse' being better than `small and dense', but high positive slopes at low $\rhonet$, indicating the rapid degradation in performance as density is reduced beyond the critical density. This is exacerbated by the fact that TIMIT with 39 \ac{MFCC}s is a dataset with low redundancy, so the effects of very low $\rhonet$ are better observed.

Fig. \ref{fig-lsvsd_timit_cifar}(b) for the \ac{MLP} portion of CIFAR-100 shows similar results as TIMIT, but on a log x-scale for more clarity. As noted in Sec. \ref{sec-trends-redundancy}, the best performance for a given $\Nnet$ occurs at an overall density less than $100\%$.  It appears that for any $\Nnet$ for CIFAR-100, peak performance occurs at around $10$--$20\%$ overall \ac{MLP} density. In experiments not shown here, we obtained similar results for the reduced redundancy net with a single convolutional layer.

\begin{comment}
Other notes:
\begin{itemize}
    \item High-level training packages (Keras, Pytorch, etc) do not have optimized sparse layers, so small dense takes less time to train, but most of the underlying acceleration packages support sparse matrix multiply operations, \ie using sparse storage formats like CSR.
    \item Does sparsity work so well with augmented datasets? - NOT TESTED
    \item Visualize features learned of different layers when using pre-defined sparsity
\end{itemize}
\end{comment}

\section{Comparison to  Other Sparse  \ac{NN} Methods}\label{sec-othersparsity}

\begin{figure}[!t]
\centering
\includegraphics[width = 1.0\linewidth]{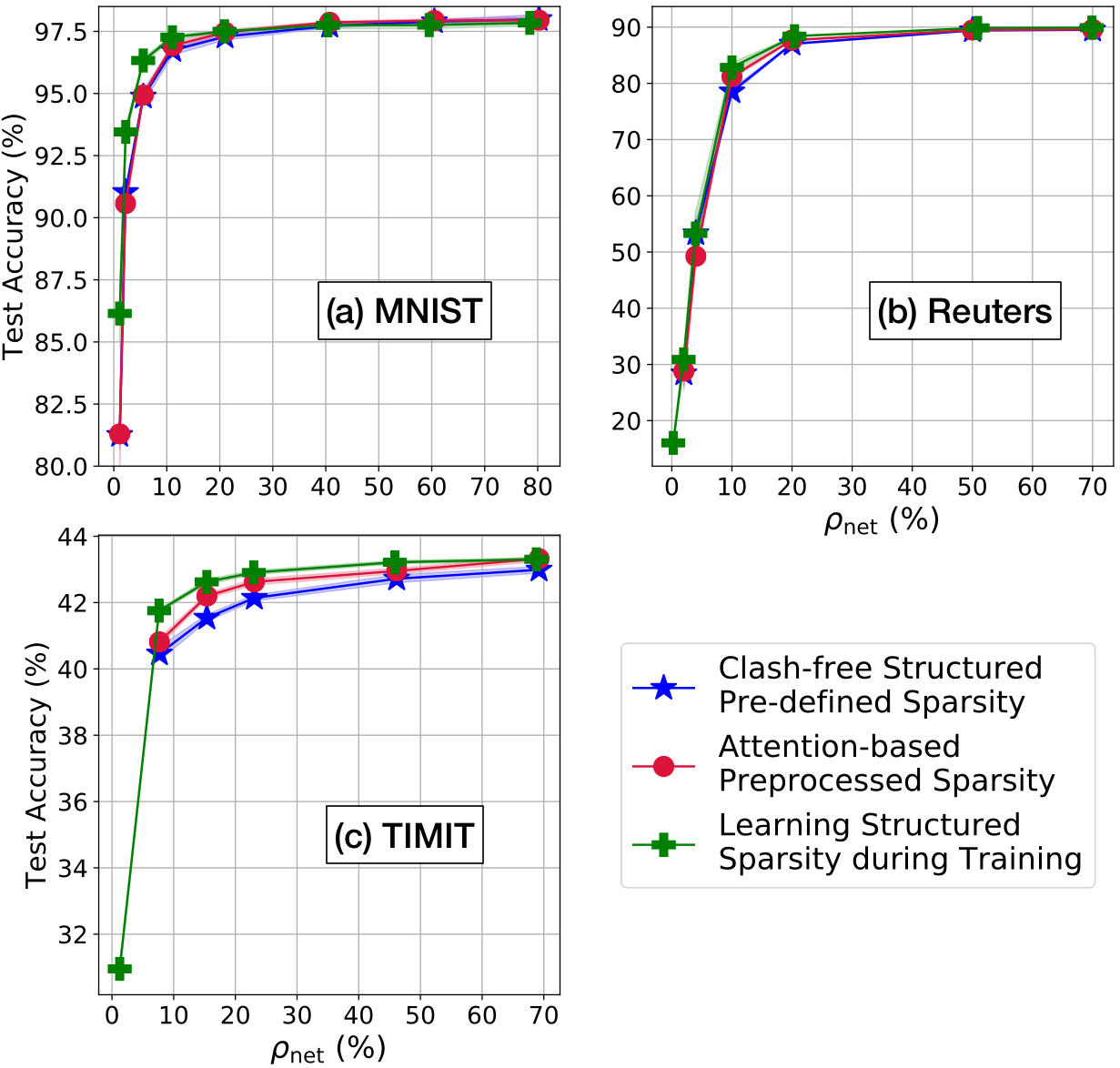}
\caption{Comparison of classification accuracy as a function of $\rhonet$ for different sparse methods on (a) MNIST with $\Nnet = (800,100,10)$, (b) Reuters with $\Nnet = (2000,50,50)$, and (c) TIMIT with $\Nnet = (39,390,39)$. We set the overall density $\rhonet$ and all individual junction densities $\rho_i$ to be approximately the same across different sparse methods. %Our proposed clash-free structured pre-defined sparse method achieves comparable performance against other sparse methods with minimal overhead on computational complexity.
}
\label{fig-other-sparse}
\end{figure}

Numerical results in  Sec.~\ref{sec-trends} showed that hardware-compatible clash-free connection patterns performed as well as structured and random pre-defined sparse connections.  In this section, we compare clash-free patterns against two sparsity approaches that are less constrained than the structured pre-defined sparsity  considered in Sec. \ref{sec-trends}.   In particular, both approaches remove the constraint of regular degree -- \ie these approaches yield sparse \ac{NN}s that have varying $\outdeg_i$ and $\indeg_i$ selected to optimize classification performance.

\subsection{Attention-based Preprocessed Sparsity}\label{sec-othersparsity-attention}
Previous works \cite{DBLP:journals/corr/BaMK14,pmlr-v37-xuc15} have applied the concept of \emph{attention} on object recognition and image captioning to achieve better performance with fewer parameters and less computation. We simplify this idea by computing the variance of input features as attention and setting the out-degree of the neurons of the input layer based on this value, Specifically, the feature variances are  quantized into three levels, and input neurons with higher attention are assigned  more connections than those with lower attention. For the neurons in latter layers, we use uniform out-degree and in-degree.

\begin{comment}
preprocessing the dataset to construct a sparse attention mask $\mathbf{p}_i \in \mathbb{N}^{N_{i-1}}$ to the neurons in layer $i-1$ according to the level of attention, such that $p_i^{(j)}$ is the number of neurons in layer $i$ to which neuron $j$ in layer $i-1$ connects.

For MNIST and Reuters, since junction 2 is more important than junction 1 as shown in Fig. \ref{fig-jnimp_mnist_rcv1_cifar}(a,c), we applied the attention-based and \ac{LSS} methods on junction 1 only. For TIMIT with 39 \ac{MFCC}s, since junction 1 is more important as shown in Fig. \ref{fig-jnimp_timit_rrcv1}(a), the attention-based method is applied with a sparse mask $\mathbf{p}_1$ calculated from the training data on junction 1, and a uniform sparse mask $\mathbf{p}_2$ on junction 2, while `learning structured sparsity' method applies the sparse-promoting penalty function to both junctions.  KUAN-WEN: these details go int th etext, not here -- ie, this is the answer to my questions in the text.
\end{comment}

\begin{comment}
a sparse attention mask $\mathbf{p}_i \in \mathbb{N}^{N_{i-1}}$ to the neurons in layer $i-1$, such that $p_i^{(j)}$ is the number of neurons in layer $i$ to which neuron $j$ in layer $i-1$ connects. 
The sparse attention mask $\mathbf{p}_i$ is constructed by analyzing and preprocessing the dataset before training. 
\end{comment}

\subsection{Learning Structured Sparsity during Training}\label{sec-othersparsity-learning}
While the method in Sec.~\ref{sec-othersparsity-attention} obtains a non-uniform neuron out-degree for the first layer, it only considers the properties of the dataset and not the learning process.  We also compared against the method of \ac{LSS} which learns a good sparse connection pattern during  training.  This method was proposed in \cite{NIPS2016_6504} and prunes the connections during training by using a  sparse-promoting penalty function as part of the objective function. Example penalty functions include L1 and L1/L2 used in Lasso\cite{Osborne99} and group-Lasso\cite{JenattonAB11}, respectively. During training, the optimizer minimizes a balancing objective comprising the loss function $l(\cdot)$\footnote{Here we emphasize that the loss function depends on all of  the trainable parameters in the network, as opposed to the output layer activations and ground truth labels as done in Sec. \ref{sec-spds-basics}. This is to emphasize that loss is a function of all of the trainable parameters and therefore the loss function can promote sparsity by driving some edge weights to zero.}, the regularizer $r(\cdot)$, and a sparse-promoting penalty function $p(\cdot)$,
\begin{equation}
\underset{\{ \bm{W}_i,\bm{b}_i\}_{i=1}^L}{\min} 
l\left( \left\{ \bm{W}_i,\bm{b}_i\right\}_{i=1}^L \right) + \lambda r\left(\left\{ \bm{W}_i\right\}_{i=1}^L\right) +  \sum_{i=1}^L \gamma_i p(\bm{W}_i)
\end{equation}
where the penalty coefficients $\{\gamma_i\}_{i=1}^L$ control the density of each junction. Increasing $\gamma_i$ decreases $\rho_i$, however, obtaining a specific value of $\rho_i$ requires experimental tuning of $\gamma_i$. In the results presented in this section, we used L1 as the element-wise sparse-promoting penalty function and L2 as the regularizer.
Note that, in contrast to the attention-based method and the structured pre-defined sparsity approach, \ac{LSS} is not a pre-defined sparsity method.  Instead training in \ac{LSS} begins with a \ac{FC} network, which means that training complexity is similar to that of a \ac{FC} \ac{NN}.  At the end of the \ac{LSS} training process, weights with absolute value below a threshold are set as zero to achieve the target density.  

\subsection{Performance comparison}\label{sec-othersparsity-comparison}
Fig. \ref{fig-other-sparse} compares performance versus $\rhonet$ of different sparse \ac{NN}s on MNIST, Reuters, and TIMIT. The individual density of each junction with the attention-based preprocessed sparse method is set to be identical to the density of each junction using clash-free pre-defined sparse method. However, the density of the nets using the \ac{LSS} method can be tuned only with the penalty coefficients. We tuned these to approximate match the density of the other methods.\footnote{This is why $\rhonet$ values of the green curves do  not perfectly align with the pre-defined sparsity curves.}

The \ac{LSS} method performs  best among all sparse methods, which is to be expected as it is the least constrained and also discovers a good sparse connection pattern during training.   However,  the performance with clash-free pre-defined sparsity is near that of the attention-based and \ac{LSS} methods -- \ie  within $2\%$ in terms of test accuracy at $\rhonet = 20\%$. We conclude that even though the clash-free patterns are highly structured and pre-defined, there is no significant performance degradation when compared to advanced methods for producing sparse models by exploiting specific properties of the dataset or learning sparse patterns during training.

\section{Conclusions and Future Work }\label{sec-conc}
In this work we  proposed a new technique for complexity reduction of neural networks -- pre-defined sparsity -- in  which a fixed sparse connection pattern is enforced prior to training and held fixed during both training and inference. We  presented a hardware architecture suited to leverage the benefits of structured pre-defined sparsity, capable of parallel and pipelined processing. The architecture can be used for both training and inference modes, and supports networks of arbitrary density, including conventional fully-connected ones. Flexibility is afforded by the degree of parallelism $\znet$, which trades hardware complexity for speed. Simple methods for clash-free memory access are presented and these methods are shown to achieve performance on par with the best known methods for obtaining sparse \ac{MLP}s.   

Using  extensive  numerical experiments, we identified trends which help in designing pre-defined sparse networks. Firstly, it is better to allocate connections in a structured manner rather than randomly. Secondly, for most datasets with high redundancy, earlier junctions can be made more sparse. Thirdly, it is better to have more neurons in the hidden layers, and then sparsify aggressively to keep the number of edges low and reduce complexity. 

As motivated in the Introduction, the rapidly growing complexity associated with modern \ac{NN}s is a major challenge. Pre-defined sparsity  is a simple method to help address this challenge, as is acceleration with custom hardware.  Interesting areas for future research include analytical  approaches to justify the  trends observed in this work and improving  our initial hardware implementation in \cite{Dey2018_Reconfig}.  It is also interesting to consider extending the methods introduced herein to convolutional layers and recurrent architectures. Finally, truly speeding the training process by orders of magnitude would allow more extensive search over \ac{NN} architectures and therefore a better understanding of the largely empirical process of \ac{NN} design.

% or


\appendices

\section{Structured Pre-Defined Sparsity Constraints}\label{appendix-spds-constraints}
In our structured pre-defined sparse network, $\rho_i$, the density of junction $i$, cannot be arbitrary, since $\rho_i = \outdeg_i/N_i = \indeg_i/N_{i-1}$, where $\outdeg_i$ and $\indeg_i$ are natural numbers satisfying the equation $N_{i-1} \outdeg_i =  N_i \indeg_i$. Therefore, the number of possible $\rho_i$ values is the same as the number of $\left(\outdeg_i,\indeg_i\right)$ values satisfying the structured pre-defined sparsity constraints:
\begin{equation}\label{eq-spds-constraints}
    \outdeg_i = \frac{N_i \indeg_i}{N_{i-1}}, ~~~
    \indeg_i \leq N_{i-1}, ~~~
    \outdeg_i, \indeg_i \in \mathbb{N}
\end{equation}
where $\mathbb{N}$ denotes the set of natural numbers.

The smallest value of $\indeg_i$ which satisfies $\outdeg_i \in \mathbb{N}$ is $N_{i-1}/\mathrm{gcd}(N_{i-1},N_i)$, and other values are its integer multiples. Since $\indeg_i$ is upper bounded by $N_{i-1}$, the total number of possible $\left(\outdeg_i,\indeg_i\right)$ is $\mathrm{gcd}(N_{i-1},N_i)$. Thus, the set of possible $\rho_i$ is
\begin{comment}
\begin{equation}
    \left\{\rho_i \bigg|\; \rho_i = \frac{k}{ \mathrm{gcd}(N_{i-1},N_i)}, k\in\{1,2,\cdots,\mathrm{gcd}(N_{i-1},N_i)\}\right\}.
\end{equation}
Which is better? 
\end{comment}

\begin{equation}
    \left\{\rho_i\in (0,1] \bigg|\; \rho_i = \frac{k}{ \mathrm{gcd}(N_{i-1},N_i)}, k\in \mathbb{N}\right\}.
\end{equation}
As a concrete example, consider a \ac{NN} with $\Nnet = (117,390,13)$. The number of possible densities of the junctions are determined by $\mathrm{gcd}(117,390) = 39$ and $\mathrm{gcd}(390,13) = 13$. Therefore, the sets of junction densities are
\begin{comment}
\begin{equation}
    \begin{aligned}
        \rho_1 &\in \left\{\frac{1}{39},\frac{2}{39},\cdots, \frac{39}{39}\right\},\\
        \rho_2 &\in \left\{\frac{1}{13},\frac{2}{13},\cdots, \frac{13}{13}\right\}.
    \end{aligned}
\end{equation}
\end{comment}
\begin{equation}
        \rho_1 \in \left\{\frac{1}{39},\frac{2}{39},\cdots, \frac{39}{39}\right\}, ~~~~~~
        \rho_2 \in \left\{\frac{1}{13},\frac{2}{13},\cdots, \frac{13}{13}\right\}.
\end{equation}

\section{Hardware Architecture Constraints}\label{appendix-hardware-constraints}
The depth of left memories in our hardware architecture is $D_i = N_{i-1}/z_i$. Thus $N_{i-1}$ should preferably be an integral multiple of $z_i$. This is not a burdening constraint since the choice of $z_i$ is independent of network parameters and depends on the capacity of the device. In the unusual case that this constraint cannot be met, the extra cells in memories can be filled with dummy values such as 0.

There are also 2 conditions placed on the $z$ values to eliminate stalls in processing: for all layers $i \in \{1,\cdots,L\}$, (i) $\left|\bm{W}_i\right|/z_i = C$, and (ii) $z_{i+1} \geq \left\lceil z_i/\indeg_i \right\rceil$. Using the definitions from Sec. \ref{sec-spds-basics}, (i) is equivalent to $z_{i+1} = z_i\outdeg_{i+1}/\indeg_i$. Then, (ii) can be equivalently written as
\begin{IEEEeqnarray}{c}\label{eq-hardware-2ndconstraint}
\outdeg_{i+1} \geq \frac{\indeg_i}{z_i} \left\lceil\frac{z_i}{\indeg_i}\right\rceil
\end{IEEEeqnarray}
which needs to be satisfied $\forall~ i \in \{1,\cdots,L-1\}$. In practice, it is desirable to design $z_i/\indeg_i$ to be an integer so that an integral number of right neurons finish processing every cycle. This simplifies hardware implementation by eliminating the need for additional storage, for example, of the intermediate activation values during \ac{FF}. In this case, \eqref{eq-hardware-2ndconstraint} reduces to $\outdeg_{i+1} \geq 1$, which is always true.

For non-integral $z_i/\indeg_i$, there are two cases. If $z_i>\indeg_i$, \eqref{eq-hardware-2ndconstraint} reduces to $\outdeg_{i+1}\geq2$. On the other hand, if $z_i<\indeg_i$, there is no bound on the right hand side of \eqref{eq-hardware-2ndconstraint}. In general, note that \eqref{eq-hardware-2ndconstraint} becomes a burdening constraint only if $\indeg_i$ is large, and $\outdeg_{i+1}$ and $z_i$ are both desired to be small. This corresponds to earlier junctions being denser than later, which is typically not desirable according to the observations in Sec. \ref{sec-trends-jndensities}, or to very limited hardware resources. We thus conclude that \eqref{eq-hardware-2ndconstraint} is not a limiting constraint in most practical cases.

\section{Clash-Free  Patterns }\label{appendix-cftypes}
Specifying $N_{i-1}$, $N_i$, $\indeg_i$ and $z_i$ for junction $i$ in a clash-free structured pre-defined sparse \ac{NN} does not uniquely define a connection pattern (unless it is \ac{FC}). This section discusses the \emph{number of possible left memory access patterns} $\countNmi$ for such a junction $i$. Note that the total number of possible memory access patterns for the complete \ac{NN} is $S_M = \prod_{i=1}^{L}{\countNmi}$.

When $z_i\geq\indeg_i$, which is expected to be true for practical cases of implementing sparse \ac{NN}s on powerful hardware devices, $\countNmi$ is also equal to the \emph{number of possible connection patterns} $\countNci$, which is the key quantity of interest. This is because if $z_i\geq\indeg_i$, at least one right neuron is completely processed in some cycle. Thus, changing the left memory access pattern will change the left neurons to which that right neuron connects, thereby changing the connection pattern. This one-to-one correspondence results in $\countNmi=\countNci$.

For the case of $z_i < \indeg_i$, a \ac{FC} junction provides an example where $\countNmi \neq \countNci$.  Specifically, in this case $\countNci=1$ as there is only one way to fully connect all neurons, but there are many clash-free memory access patterns, as shown in the following equations \eqref{eq-type1cf-count}-\eqref{eq-type3cf-count}.

\begin{figure}[!t]
\centering
\includegraphics[width = 0.6\linewidth]{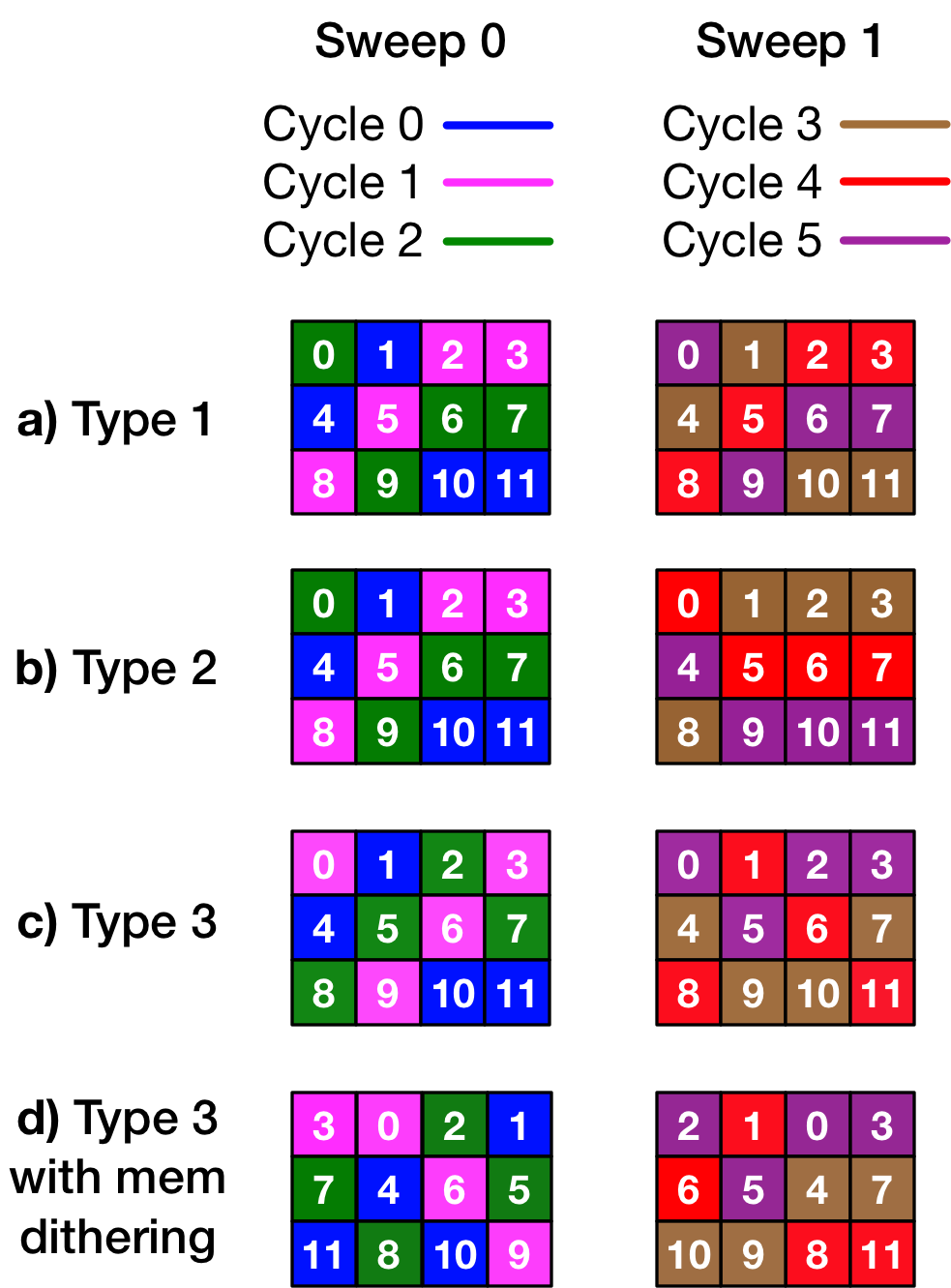}
\caption{(a-c) Various types of clash-freedom, and (d) memory dithering for type 3, using the same left neuronal structure from Fig. \ref{fig-memaccesses} as an example. The grids represent different access patterns for the same memory bank. The number in each cell represents the left neuron number whose parameter is stored in that cell. Cells sharing the same color are read in the same cycle.}
\label{fig-clashfreedom}
\end{figure}

We now discuss various types of clash-freedom, and $\countNmi$ arising from each:
\begin{itemize}
    \item Type 1: This is as described in Sec. \ref{sec-hardware-clashfreedom}, and recapitulated in Fig. \ref{fig-clashfreedom}(a). $\countNmi$ is the number of ways of designing $\bm{\phi}_i$, \ie
    \begin{IEEEeqnarray}{c}\label{eq-type1cf-count}
    \countNmi = {D_i}^{z_i}
    \end{IEEEeqnarray}
    %As an example, consider the \ac{NN}s from Fig. \ref{fig-pdsexample}(c) with $\Nnet=(800,100,10)$. Choosing $\znet = (100,25)$, which is similar to our implementation in \cite{Dey2018_Reconfig}, we get $\countNmi$ on the order of $10^{105}$, which we believe is sufficient for most applications. The $\countNmi$ value obtained from eq. \eqref{eq-type1cf-count} can be too small for pathological values of $\znet$. Referring to it as \emph{type 1 clash-freedom}, here we describe other types of clash-free memory accesses which increase $\countNmi$ at a greater hardware cost. These are shown in Fig. \ref{fig-clashfreedom}.
    
    \item Type 2 (implemented in our earlier work \cite{Dey2018_Reconfig}): In this technique, a new $\bm{\phi}_i$ is defined for every sweep. Considering the example in Fig. \ref{fig-clashfreedom}(b), $\bm{\phi}_i=(1,0,2,2)$ for sweep 0, but $(2,0,0,0)$ for sweep 1. There will be $\outdeg_i$ different $\bm{\phi}_i$ vectors for each junction, resulting in:
    \begin{IEEEeqnarray}{c}\label{eq-type2cf-count}
    \countNmi = {D_i}^{z_i\outdeg_i}
    \end{IEEEeqnarray}
    
    \item Type 3: In this technique, the constraint of cyclically accessing the left memories is also eliminated. Instead, any cycle can access any cell from each of the memories. This means that storing $\bm{\phi}_i$ is not enough, the entire sequence of memory accesses needs to be stored as a matrix $\bm{\Phi}_i \in \{0,1,\cdots,D_i-1\}^{D_i\times z_i}$. In Fig. \ref{fig-clashfreedom}(c) for example, $\bm{\Phi}_i = ((1,0,2,2),(0,2,1,0),(2,1,0,1))$ for sweep 0. Every sweep would also have a different $\bm{\Phi}_i$, resulting in:
    \begin{IEEEeqnarray}{c}\label{eq-type3cf-count}
    \countNmi = \left(D_i!\right)^{z_i\outdeg_i}
    \end{IEEEeqnarray}
    
    %The \emph{most general kind} of clash-freedom resulting in the biggest $\countNmi$ arises from ignoring the effect of sweeps. This means that each address in each left memory of junction $i$ can be accessed in any cycle, as long as it is accessed a total of $\outdeg_i$ times during the junction cycle (consisting of $C=D_i\outdeg_i$ cycles). However, this may lead to connectivity patterns where a pair of neurons in adjacent layers have more than 1 connection between them, which is invalid due to redundancy. Hence we skip the details, but state the resulting count of possible \ac{NN}s:
    %\begin{IEEEeqnarray}{c}\label{eq-generalcf-count}
    %\countNmi = \left( \frac{C!}{\left(\outdeg_i!\right)^{D_i}} \right)^{z_i}
    %\end{IEEEeqnarray}
\end{itemize}

A technique that can be applied to all the types of clash-freedom is \emph{memory dithering}, which is a permutation of the $z_i$ memories (\ie the columns) in a bank. This permutation can change every sweep, as shown in Fig. \ref{fig-clashfreedom}(d). Memory dithering incurs an additional address computation storage cost because of the $z_i$ permutation, but increases $\countNmi$ by a factor $K_i$. If $\indeg_i/z_i$ is an integer, an integral number of cycles are required to process each right neuron. Since a cycle accesses all memories, dithering has no effect and $K_i=1$. On the other hand, if $z_i/\indeg_i$ is an integer greater than 1, the effects of dithering on connectivity patterns are only observed when switching from one right neuron to the next within a cycle. This results in
\begin{IEEEeqnarray}{c}
K_i = \left(\frac{z_i!}{{\indeg_i!}^{\frac{z_i}{\indeg_i}}}\right)^{\outdeg_i}
\end{IEEEeqnarray}
for types 2 and 3, and the $\outdeg_i$ exponent is omitted for type 1 since the access pattern does not change across sweeps.

When either of $z_i$ or $\indeg_i$ does not perfectly divide the other, an exact value of $K_i$ is hard to arrive at since some proper or improper fraction of right neurons are processed every cycle. In such cases, $K_i$ is upper-bounded by ${\left(z!\right)}^{\outdeg_i}$.
\begin{comment}
Thus the maximum number of valid \ac{NN} connectivity patterns arises from type 3 with memory dithering, given as:
\begin{IEEEeqnarray}{c}\label{eq-type3cfmd-count}
\countNmi = \prod_{i=1}^{L}{K_i\left(D_i!\right)^{z_i\outdeg_i}}
\end{IEEEeqnarray}
\end{comment}

Table \ref{table-clashfreedom-comparison} compares the count of possible left memory access patterns and associated storage cost for computing memory addresses for types 1--3, with and without memory dither. The junction used is the same as in Fig. \ref{fig-memaccesses}, except $N_i$ is raised to 12 such that $\indeg_i$ becomes 2 and allows us to better show the effects of memory dithering.

\begin{table}[!t]
\renewcommand{\arraystretch}{1.2}
\caption{Comparison of Clash-Free Methods for a Single Junction $i$ with $(N_{i-1},N_i,\outdeg_i,\indeg_i,z_i) = (12,12,2,2,4)$}
\label{table-clashfreedom-comparison}
\centering
\begin{tabular}{|c|c|c|c|}
\hline
\multirow{2}{*}{Type} & Memory & \multirow{2}{*}{$\countNmi$} & Storage Cost to Compute \\
& Dithering & & Memory Addresses \\
\hline
\multirow{2}{*}{1} & No & 81 & $z_i = 4$ \\
\cline{2-4}
& Yes & 486 & $2z_i = 8$ \\
\hline
\multirow{2}{*}{2} & No & 6561 & $z_i\outdeg_i = 8$ \\
\cline{2-4}
& Yes & 236k & $2z_i\outdeg_i = 16$ \\
\hline
\multirow{2}{*}{3} & No & 1.68M & $N_{i-1}\outdeg_i = 24$ \\
\cline{2-4}
& Yes & 60M & $\left(N_{i-1}+z_i\right)\outdeg_i = 32$ \\
\hline
\end{tabular}
\end{table}

%\section*{Acknowledgment}


\bibliographystyle{IEEEtran}
\bibliography{IEEEabrv,aaa_main}

\end{document}